\pdfoutput=1
\documentclass[11pt]{article}

\usepackage[]{ACL2023}

\usepackage{times}
\usepackage{latexsym}
\usepackage{graphicx}

\usepackage[T1]{fontenc}
\usepackage[utf8]{inputenc}
\usepackage{microtype}
\usepackage{inconsolata}
\usepackage{caption}
\usepackage{subcaption}
\usepackage{booktabs}
\usepackage{xcolor,soul}
\usepackage{framed}

\title{\dataset{}: A Dataset for Document-Level Simplification of Wikipedia Pages}

\iffalse
    \newcommand{\wk}[1]{}
    \newcommand{\pl}[1]{}
    \newcommand{\jw}[1]{}
    \newcommand{\jv}[1]{}
    \newcommand{\sj}[1]{}
\else
    \newcommand{\wk}[1]{\textcolor{orange}{\textbf{WK:} #1}}
    \newcommand{\pl}[1]{\textcolor{blue}{\textbf{PL:} #1}}
    \newcommand{\jw}[1]{\textcolor{yellow}{\textbf{JW:} #1}}
    \newcommand{\jv}[1]{\textcolor{green}{\textbf{JV:} #1}}
    \newcommand{\sj}[1]{\textcolor{red}{\textbf{SJ:} #1}}
\fi
\newcommand*{\affaddr}[1]{#1}

\author{
  \quad \textbf{Philippe Laban}
  \quad \textbf{Jesse Vig}
  \quad \textbf{Wojciech Kryscinski} \\
  \quad \textbf{Shafiq Joty}
  \quad \textbf{Caiming Xiong}
  \quad \textbf{Chien-Sheng Jason Wu} \\
  \affaddr{Salesforce AI} \\
  \{plaban, jvig, wojciech.kryscinski, sjoty, cxiong, wu.jason\}@salesforce.com \\
}

\begin{document}
\newcommand\Tstrut{\rule{0pt}{2.8ex}}       
\newcommand\Bstrut{\rule[-1.4ex]{0pt}{0pt}} 
\newcommand{\TBstrut}{\Tstrut\Bstrut} 

\newcommand{\dataset}{\textsc{SWiPE}}

\definecolor{colordel}{HTML}{FECACA}
\definecolor{colorins}{HTML}{A1D8FC}

\definecolor{colorlex}{HTML}{E69F00}
\definecolor{colorsyn}{HTML}{57B4E9}
\definecolor{colordis}{HTML}{019E73}
\definecolor{colorsem}{HTML}{CC79A7}
\definecolor{colornon}{HTML}{ABABAB}

\DeclareRobustCommand{\hlred}[1]{{\sethlcolor{colordel}\hl{#1}}}
\DeclareRobustCommand{\hlblue}[1]{{\sethlcolor{colorins}\hl{#1}}}

\maketitle

\begin{abstract}
Text simplification research has mostly focused on sentence-level simplification, even though many desirable edits---such as adding relevant background information or reordering content---may require document-level context.
Prior work has also predominantly framed simplification as a single-step, input-to-output task, only implicitly modeling the fine-grained, span-level edits that elucidate the simplification process.
To address both gaps, we introduce the \dataset{} dataset, which reconstructs the \textit{document-level}  \textit{editing} process from English Wikipedia (EW) articles to paired Simple Wikipedia (SEW) articles. 
In contrast to prior work, \dataset{} leverages the entire revision history when pairing pages in order to better identify simplification edits. We work with Wikipedia editors to annotate 5,000 EW-SEW document pairs, labeling more than 40,000 edits with proposed 19 categories.
To scale our efforts, we propose several models to automatically label edits, achieving an F-1 score of up to 70.6, indicating that this is a tractable but challenging NLU task. 
Finally, we categorize the edits produced by several simplification models and find that \dataset{}-trained models generate more complex edits while reducing unwanted edits.
\end{abstract}

\section{Introduction}

\begin{figure}
    \centering
    \includegraphics[width=0.5\textwidth]{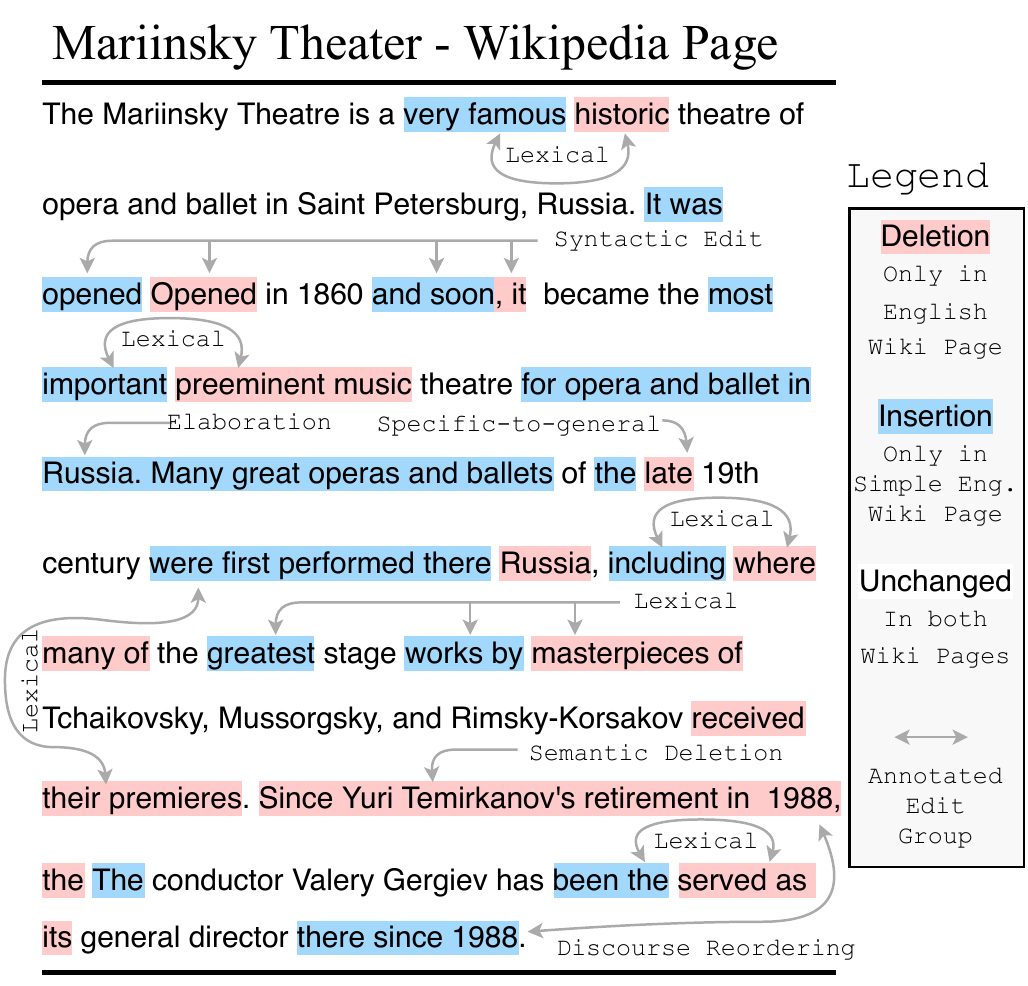}
    \caption{Sample from \dataset{}, a Wikipedia-based dataset for document-level simplification. Many edits in \dataset{} require document-level context.}
    \label{fig:illustrative_example}
\end{figure}

Text simplification (TS) aims to make complex documents accessible to larger audiences by lowering the barrier of reading for children, non-native speakers, and novice readers in technical domains.
TS has primarily been approached in a sentence-level sequence-to-sequence (seq2seq) manner, following the methodology of mature NLG tasks such as machine translation. Prior work framed at the sentence level has focused on simplification edits that occur within sentence units, such as lexical replacements \cite{glavavs2015simplifying} and sentence splitting \cite{Narayan2015UnsupervisedSS,sulem2018bleu}. Yet, many simplification operations, such as background elaborations \cite{srikanth2020elaborative} or content reordering \cite{zhong2020discourse} require document-level context.

A major roadblock to advances in document-level simplification has been the lack of large-scale and high-quality datasets.
The two most popular sources of data for the English language are either the news-based Newsela which is not available publicly or the combination of English Wikipedia (EW) and Simple English Wikipedia (SEW)\footnote{\url{https://simple.wikipedia.org}}, which is large-scale but requires non-trivial processing to align Wikipedia articles with their simplified versions~\cite{jiang2020neural}. The alignment task has predominantly been framed as finding pairs of semantically similar sentences within the latest revisions of EW and SEW pages.

Our first contribution is to adapt the Wikipedia content alignment task to document-level granularity. We explore the entire \textit{revision history} of Wikipedia pages and match individual revisions of SEW pages with best-aligned EW revisions, rather than rely on the most recent revisions which might yield factually misaligned pairs due to outdated information. By applying our alignment method to the entire revision history of SEW -- and processing two orders of magnitude more content --  we create the \dataset{} dataset, a high-quality and large-scale document-level simplification dataset. \dataset{} consists of 145,161 document pairs, which we processed into an alignment sequence composed of three operations: \textit{unchanged text}, \textit{insertion}, and \textit{deletion}. Figure~\ref{fig:illustrative_example} provides an illustrative alignment sequence of a \dataset{} sample.

Our second contribution is a comprehensive analysis of edits that occur in \dataset{}. We propose a 19-category edit taxonomy based on prior work and expanded for document-level edits. The categories are organized into four coarse-grained classes representing simplification objectives: Lexical, Syntactic, Semantic, and Discourse-level edits. We collaborate with active SEW editors to annotate 5,000+ alignment sequences of \dataset{}. The collected annotations of around 40,000 edits reveal that all four edit classes are prevalent in \dataset{} (each occurs in at least 40\% of annotated documents). Document-level context is required for at least 43\% of edits, and diverse edits often co-occur within documents, as SEW editors combine editing strategies when producing SEW pages.

Our third contribution is to propose models that can automatically identify edit categories and models that generate document-level simplified text. For the task of edit identification, our best model achieves a categorization F-1 score of 70.6, leaving room for future improvement. When analyzing simplification models based on the edits they produce, we find that \dataset{}-trained models can produce more complex edits than prior work while generating fewer undesirable edits that potentially introduce factually incorrect content. We release the \dataset{} data, the models, and experimental code publicly\footnote{\url{https://github.com/Salesforce/simplification}}.

\section{Related Work}

\subsection{Simplification Datasets}

Simple Wikipedia was leveraged by prior work to create some of the first large-scale simplification resources, such as PWKP \cite{zhu2010monolingual} and SEW \cite{coster2011simple}, which popularized the field framed on sentence-level simplification. Subsequent work found shortcomings in initial datasets due to low-quality alignment \cite{xu2015problems}, and three main avenues for improvement were proposed. First, some work proposed to favor higher quality data sources such as Newsela \cite{xu2015problems,srikanth2021elaborative}. However, Newsela is only available under a restrictive license, which has limited its accessibility within the research community. Second, manual annotation of smaller-scale but higher-quality evaluation sets can complement existing resources, such as HSplit \cite{sulem2018bleu}, TurkCorpus \cite{xu2016optimizing}, and ASSET \cite{alva2020asset}. Finally, more advanced alignment methods were proposed to improve the automatic creation of Wikipedia-based datasets, creating Wiki-Auto \cite{jiang2020neural} and CATS \cite{vstajner2018cats}.

Recent work has explored simplification beyond sentence-level granularity, with some methods focused on the paragraph level \cite{devaraj2021paragraph,laban2021keep}. The D-Wikipedia dataset \cite{sun2021document} is the closest in format to \dataset{}, but analysis in Section~\ref{section:dwiki_comparison} reveals that it is of limited quality due to a lack of filtering. With \dataset{}, we extend prior work by implementing an advanced automatic alignment method to create a large-scale dataset for document-level simplification.


\subsection{Categorizing Simplification Edits}

Given a simplification dataset, automatic alignment methods enable the extraction of atomic edits that simplify the complex text. Prior work has analyzed such edits to gain insights and compare datasets. The most common analysis revolves around measuring the frequency of different editing operations (i.e. insertions, deletions, replacements) \cite{coster2011simple, vasquez2021role}. Some work has proposed annotating the operations with linguistically motivated categories that give a reason for the edit. Since most simplification resources are at the sentence granularity, edit categorizations have focused on lexical and syntactic phenomena that frequently occur within individual sentences \cite{aluisio2008corpus,scarton2018learning,cardon2022linguistic}.

Some work has leveraged Newsela to study edits that require document-level context, such as elaborations \cite{srikanth2020elaborative} and content selection \cite{zhong2020discourse}. Other works such as arxivEdits\cite{jiang2022arxivedits}, EditEval\cite{dwivedi2022editeval} or PEER\cite{schick2022peer} have studied the general problem of document editing, and have either not considered simplification edits or grouped all simplification edits within a single category. Edit categories used in \dataset{} are based on existing categorization and expanded with edits less frequently, studied such as discourse and semantic edits that require document-level context.

\subsection{Wikipedia Revision History}

Wikipedia revision history has been used in NLP resources, from automatic grammatical error correction \cite{boyd2018using,max2010mining}, to vandalism detection \cite{chin2010detecting,heindorf2015towards}, paraphrase generation \cite{nelken2008mining,dutrey2010local} or fact verification \cite{schuster2021get}. With \dataset{}, we show that Wikipedia's revision history in conjunction with advanced alignment methods can be a powerful tool to create simplification datasets.

\section{Creating \dataset}

\begin{figure}
    \centering
    \includegraphics[width=0.48\textwidth]{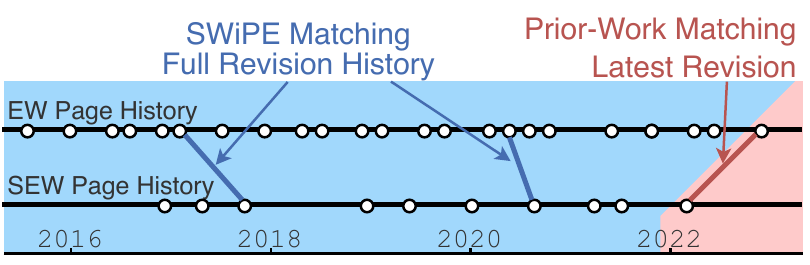}
    \caption{\dataset{} matching considers the entire revision history of pages, enabling better page alignment.}
    \label{fig:revision_history}
\end{figure}

\subsection{Page Matching}
To create a simplification dataset based on Wikipedia, pages from EW must be matched with their counterpart simplified pages in SEW. We follow prior work and leverage Wikidata \cite{jiang2020neural}, Wikimedia's knowledge base, to extract Wikidata entries with both EW and SEW Wikipedia pages, and obtain a total of 226,861 page pairs, which form the basis for our dataset.

\subsection{Revision Matching}
By design, each Wikipedia page is a living document that is continuously updated and revised. When an editor creates a SEW page, it is common practice to select a particular revision of the corresponding EW page as a starting point and introduce a series of simplifying edits.

Most existing Wikipedia-based simplification datasets rely on matching the latest revisions of page pairs at the time of dataset creation, overlooking page revision history. Considering that EW pages are typically updated more frequently than SEW pages, such approaches might lead to misalignment in the created datasets, thus lowering the data quality. 
In this work, we leverage the full revision history of both the EW and SEW pages with the goal of obtaining higher-quality examples of document-level simplification. We propose the task of automatic revision matching, illustrated in Figure~\ref{fig:revision_history}.

For the 226,861 page pairs, we obtain the entire revision history of the EW and SEW pages and extract up to 200 full-text revisions using Wikipedia's API. We obtain 22 million revisions: on average 94 revisions per EW page, and 4 per SEW page.
The matching process consists of finding the EW revision that aligns best with each SEW revision. If a SEW page has multiple revisions, we include several revisions in the dataset, as long as the SEW revisions differ significantly and match distinct EW revisions (i.e., Levenshtein similarity $\leq$0.3).

We manually annotated 2,000 revision pairs with an alignment label (0/1) and conducted an exploratory study of several baseline models, with full details in Appendix~\ref{appendix:revision_matching}. Based on the findings, we select the NLI-based \texttt{SummaC} model~\cite{laban2022summac}, which was originally proposed for inconsistency detection in summarization, as the final alignment model. The model achieved a strong performance of 91.5 recall and 84.2 F-1 on a held-out test set.

It is possible for SEW revisions to match none of its paired EW revisions if the SummaC model predicts that all pairs are unaligned. This occurs frequently, for example when a SEW page is written without being based on the relevant EW page. In total, matches occur for 133,744 page pairs, leading to a total of 145,161 revision-pair matches.


In Section~\ref{section:annotation}, Wikipedia editors participating in \dataset{}'s annotation could flag samples they deemed unaligned. Of the roughly 5,000 annotated samples, just 4\% were flagged as unaligned, validating the high precision of the matching process.

\subsection{\dataset{} Statistics}

We focus the dataset on the introduction section of each Wikipedia page, as prior work has shown that including all sections leads to a large imbalance in terms of length \cite{xu2015problems}.

The average compression ratio from EW to SEW page in \dataset{} document pairs is 0.87, suggesting that SEW pages are not significantly shorter than their EW matches. In fact, 26\% of document pairs have a compression ratio larger than 1, indicating that is not infrequent for the simplification of a document to be longer than the original document.

\subsection{Comparison with Prior Work}
\label{section:dwiki_comparison}

We perform an analysis of D-Wikipedia, an existing document-level simplification dataset that was created without considering the revision history and without filtering pages based on alignment quality.

We find that of the 132,546 samples in the training portion of D-Wikipedia, only 49,379 (or 37\%) pass the alignment filtering we applied to create \dataset{}. Models trained on noisy datasets due to low-quality alignment have been shown to exhibit undesirable behavior, such as hallucinating facts in summarization \cite{maynez2020faithfulness,kryscinski2020evaluating}, which is likely to occur in simplification as well. This analysis illustrates that matching revisions from the entire revision history is an essential step in creating large-scale, high-quality simplification datasets based on Wikipedia. 

\section{Edit-Level Annotation}
\label{section:annotation}

In upcoming sections, we use the term \textit{document} to refer to a particular page version. Given two matched documents, they can be represented as a single \textit{alignment sequence} using a string-alignment algorithm such as Levenshtein \cite{Levenshtein1966BinaryCC}. An alignment sequence consists of a series of three operations: \textit{unchanged text}, \textit{inserted text}, and \textit{removed text}, as illustrated in Figure~\ref{fig:illustrative_example}. To understand the types of edits that occur in \dataset{}, we collaborated with Simple Wikipedia editors to annotate a subset of the dataset.

\subsection{Annotation Procedure Definition}

The annotation procedure of a document pair consists of selecting groups of edit operations (i.e., insertions and deletions) and assigning them to an edit category from a predefined list. A document pair is considered fully annotated once each edit operation is assigned to at least one edit group.

Edit groups can consist of a single edit operation (e.g. the Background Elaboration in Figure~\ref{fig:illustrative_example}), or multiple operations (e.g. four operations for the syntactic edit). Operations can be part of multiple groups, which enables group overlap (e.g., the second to last deletion in Figure~\ref{fig:illustrative_example} is part of Semantic Deletion and Discourse Reordering groups).

We choose to treat each operation as atomic and do not allow the annotator to manually split edit operations further. Although this could be limiting for longer edits, we believe this sets a common ground for annotation, as work in extractive QA has shown that disagreement of span boundaries affects dataset quality \cite{Rajpurkar2016SQuAD1Q}.
Analysis in Section~\ref{section:annotation_analysis} examines the prevalence of overlap and interleaving of edits in the dataset.

\subsection{Edit Categorization}
\label{section:categorization}
\begin{table}[]
    \resizebox{0.5\textwidth}{!}{%
    \begin{tabular}{lrrrrrr}
    \textbf{Edit Category} & \textbf{N} & \textbf{\%$\exists$} & \textbf{\#O} & \textbf{\%I} & \textbf{\%D} & \textbf{\%I+D} \\
    \cmidrule(r){1-1} \cmidrule(r){2-4} \cmidrule(r){5-7}
    {\Large \color{colorlex} $\bullet$} Lexical Edit & 6798 & 61.7 & 2.1 & 0.3 & 0.2 & \textbf{99.5} \\
    {\Large \color{colorlex} $\bullet$} Entity Edit & 359 & 6.4 & 1.5 & 7.2 & \textbf{57.1} & 35.7 \\
    \cmidrule(r){1-1} \cmidrule(r){2-4} \cmidrule(r){5-7}
    {\Large \color{colorsyn} $\bullet$} Sentence Split & 3010 & 43.8 & 2.3 & 42.0 & 0.3 & \textbf{57.7} \\
    {\Large \color{colorsyn} $\bullet$} Sentence Fusion & 334 & 6.0 & 2.4 & 5.7 & 29.0 & \textbf{65.3} \\
    {\Large \color{colorsyn} $\bullet$} Syntactic Deletion & 1889 & 28.1 & 1.1 & 0.2 & \textbf{98.1} & 1.7 \\
    {\Large \color{colorsyn} $\bullet$} Syntactic Generic & 2615 & 36.2 & 1.5 & 31.1 & 27.8 & \textbf{42.6} \\
    \cmidrule(r){1-1} \cmidrule(r){2-4} \cmidrule(r){5-7}
    {\Large \color{colordis} $\bullet$} Reordering & 2379 & 34.6 & 2.5 & 0.6 & 0.4 & \textbf{99.0} \\
    {\Large \color{colordis} $\bullet$} Anaphora Resolut. & 302 & 5.4 & 1.8 & 21.9 & 7.9 & \textbf{70.2} \\
    {\Large \color{colordis} $\bullet$} Anaphora Insert. & 362 & 6.4 & 1.8 & 20.4 & 0.6 & \textbf{79.0} \\
    \cmidrule(r){1-1} \cmidrule(r){2-4} \cmidrule(r){5-7}
    {\Large \color{colorsem} $\bullet$} Elaboration - Bkgrd & 805 & 12.9 & 1.4 & \textbf{93.2} & 0.4 & 6.5 \\
    {\Large \color{colorsem} $\bullet$} Elaboration - Exple & 139 & 2.4 & 1.5 & \textbf{95.7} & 0.0 & 4.3 \\
    {\Large \color{colorsem} $\bullet$} Elaboration - Generic & 3195 & 36.0 & 1.2 & \textbf{95.9} & 1.1 & 2.9 \\
    {\Large \color{colorsem} $\bullet$} Semantic Deletion & 12928 & 76.8 & 2.0 & 0.4 & \textbf{98.8} & 0.8 \\
    {\Large \color{colorsem} $\bullet$} Specific-to-General & 332 & 5.7 & 2.1 & 0.0 & 6.9 & \textbf{93.1} \\
    \cmidrule(r){1-1} \cmidrule(r){2-4} \cmidrule(r){5-7}
    {\Large \color{colornon} $\bullet$} Format & 2688 & 35.3 & 1.9 & 9.7 & 10.5 & \textbf{79.7} \\
    {\Large \color{colornon} $\bullet$} Noise Deletion & 693 & 10.6 & 1.6 & 2.2 & \textbf{93.7} & 4.2 \\
    {\Large \color{colornon} $\bullet$} Fact Correction & 290 & 5.0 & 2.3 & 4.5 & 2.8 & \textbf{92.8} \\
    {\Large \color{colornon} $\bullet$} Extraneous Info & 3028 & 36.5 & 2.2 & \textbf{99.4} & 0.1 & 0.5 \\
    {\Large \color{colornon} $\bullet$} Miscellaneous & 241 & 3.6 & 1.7 & \textbf{68.9} & 1.7 & 29.5 \\
    \bottomrule{}
    \end{tabular}
    }
    \caption{Edit categories in \dataset{}. Categories belong to five classes: {\Large \color{colorlex} $\bullet$} lexical, {\Large \color{colorsyn} $\bullet$} syntactic, {\Large \color{colordis} $\bullet$} discourse, {\Large \color{colorsem} $\bullet$} semantic, and {\Large \color{colornon} $\bullet$} non-simpl. \textbf{N}: number of annotated instances, \textbf{\%$\exists$}: percentage of documents with the edit, \textbf{\#O}: average group size, \textbf{\%I, \%D, \%I+D}: distribution over operation type (insert-only, delete-only, replace)}
    \label{table:edit_categories}
\end{table}

Edit categories were formalized by combining prior-work categorizations \cite{siddharthan2014survey,cardon2022linguistic}. Three of the authors then iteratively annotated common samples in batches of 10-20 and introduced new categories specific to document-level simplification that did not arise in sentence-level-based work. We measured inter-annotator agreement at each iteration using Fleiss' Kappa and halted once no new category was introduced and the agreement level was above 0.7.

The final categories are organized into four higher-level classes: \textbf{Lexical} edits that simplify word units; \textbf{Syntactic} edits that simplify sentence structure; \textbf{Discourse} edits that deal with multi-sentence simplification; \textbf{Semantic} edits that add or remove information within the document. An additional class handles all \textbf{Non-Simplification} edits. Each class is subdivided into categories, for a total of 19 categories. For example, the Syntactic class contains \textit{Sentence Splitting}, \textit{Sentence Fusion}, \textit{Syntactic Deletion,} and \textit{Syntactic Generic}. Classes and edit categories are listed in Table~\ref{table:edit_categories}.
A document with a definition and a canonical example of each category was prepared and later used to onboard annotators (Appendix~\ref{appendix:category_definitions}).

\subsection{Annotation Collaboration}
\begin{table}[]
    \centering
    \resizebox{0.42\textwidth}{!}{%
    \begin{tabular}{lcccc}
     & \multicolumn{3}{c}{\begin{tabular}[c]{@{}c@{}}\textbf{In-Domain}\\ 71,702 cats.\end{tabular}} & \begin{tabular}[c]{@{}c@{}}\textbf{OOD}\\ 3 cats.\end{tabular} \\
     \cmidrule(r){2-4} \cmidrule(r){5-5}
     & \textbf{Train} & \textbf{Valid} & \textbf{Test} & \textbf{Test} \\
     \cmidrule(r){1-1} \cmidrule(r){2-4} \cmidrule(r){5-5}
    Manually Annotated & 3,861 & 484 & 484 & 377 \\
    Silver Annotated & 126k & 7k & 7k & - \\
    \bottomrule
    \end{tabular}
    }
    \caption{Number of docs in each split of \dataset{}}
    \label{table:data_statistics}
\end{table}

We collaborated with active Simple Wikipedia editors to annotate \dataset{}. We contacted the 50 all-time top editors of Simple English Wikipedia on their public Wikipedia talk pages\footnote{\url{https://en.wikipedia.org/wiki/Help:Talk_pages}} with a high-level description of our project and prompted them to participate for a remuneration of US\$25/hour.

In total, six SEW editors replied to the initial message. They were given a 1-hour onboarding task to attentively read through edit category definitions and annotate ten warm-up documents spanning all edit categories. The SEW editors were invited to join a Slack channel to discuss borderline and unclear examples. Upon completion, the authors of the paper reviewed the warm-up document annotations, and annotation errors were discussed with the participants before proceeding with the actual annotation.

In total, 3 SEW editors successfully completed the onboarding, and we recruited an additional editor with a linguistic background recommended by one of the editors (not an active SEW editor). Over a period of two months, annotators identified edits in over 5,000 unique document alignment sequences. During the annotation process, annotations were periodically reviewed and feedback was given to annotators. Annotating a single sequence took an average of 1.3 minutes, and the annotation effort cost approximately US\$2,500. 

To inspect annotation quality, 329 alignment sequences were annotated by several annotators. The agreement level is measured using Fleiss' Kappa and averages 0.62 for the five category classes, indicating moderate agreement. Appendix~\ref{appendix:categories_extras} provides category-specific agreement levels, which vary across categories. Even with clear edit category definitions, there remains subjectivity in the annotation procedure.

Wikipedia categories are assigned to pages that identify page themes (e.g., Technology). In total, 71,705 Wikipedia categories appear in \dataset{}. We set aside three categories -- Materials, Economics, and Desserts -- containing 377 pairs, which we fully annotated as a more challenging out-of-domain (OOD) test set. The rest of the annotation was performed on a random sample of all other categories. Table~\ref{table:data_statistics} summarizes the number of documents in each portion of the dataset.

\subsection{Annotation Analysis}
\label{section:annotation_analysis}

\begin{figure}
    \centering
    \begin{subfigure}[b]{0.49\textwidth}
        \includegraphics[width=\textwidth]{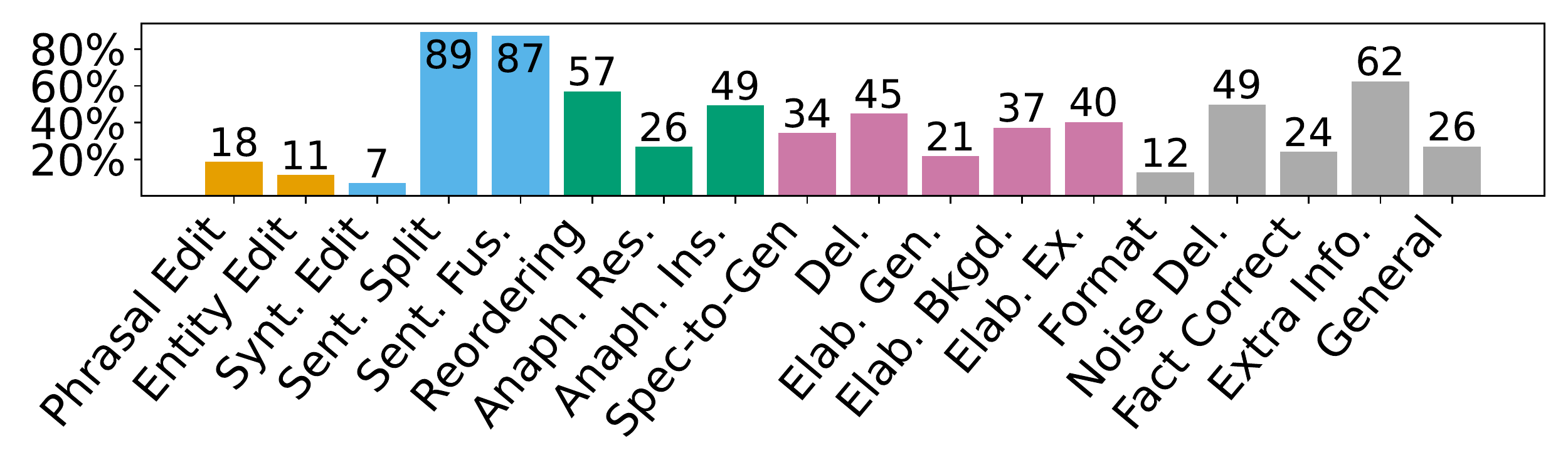}
        \vspace{-20pt}
        \caption[]{Percentage of edits that cross sentence boundaries}
        \vspace{10pt}
        \label{fig:analysis_multi_sent}
    \end{subfigure}
    \hfill
    \hfill
    \begin{subfigure}[b]{0.49\textwidth}
        \includegraphics[width=\textwidth]{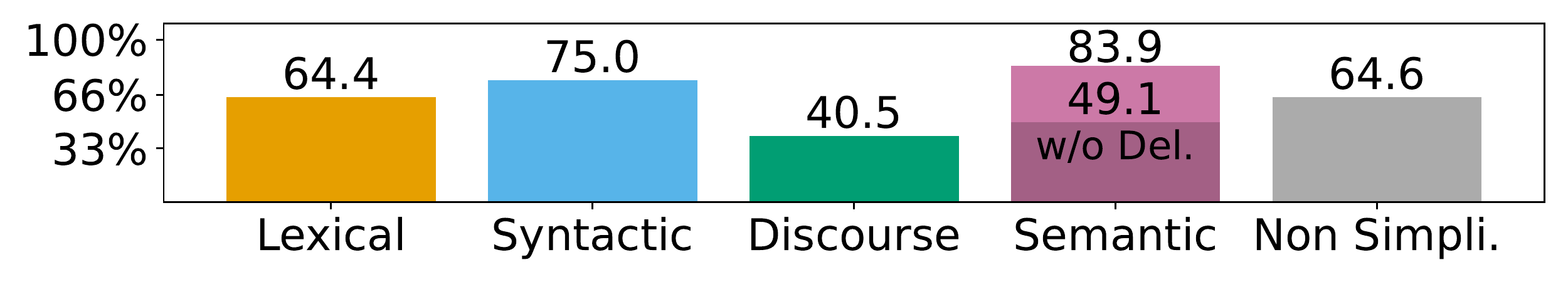}
        \vspace{-20pt}
        \caption[]{Percentage of docs that include edits from category group}
        \vspace{10pt}
        \label{fig:analysis_frequency}
    \end{subfigure}
    \hfill
    \hfill
    \begin{subfigure}[b]{0.49\textwidth}
        \includegraphics[width=\textwidth]{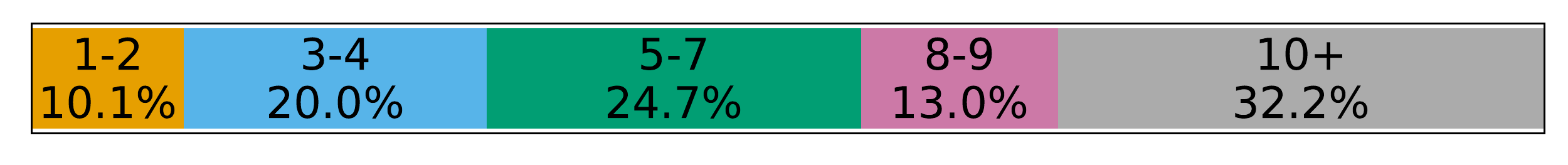}
        \vspace{-20pt}
        \caption[]{Distribution of the number of edits per document}
        \label{fig:analysis_num_groups}
        \vspace{10pt}
    \end{subfigure}
    \hfill
    \hfill
    \begin{subfigure}[b]{0.49\textwidth}
        \includegraphics[width=\textwidth]{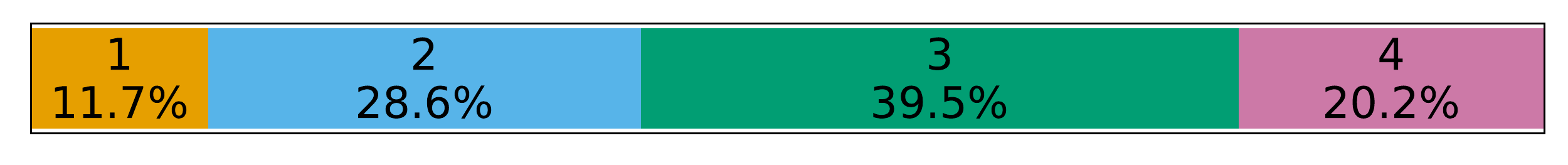}
        \vspace{-20pt}
        \caption[]{Distribution of distinct category classes within documents}
        \label{fig:analysis_joint_groups}
    \end{subfigure}

    \caption[]{Summary of annotation analysis}
    \label{fig:annotation_analysis}
\end{figure}

Figure~\ref{fig:annotation_analysis} summarizes \dataset{} annotations statistics. In Figure~\ref{fig:analysis_multi_sent}, we break down the percentage of edits that cross sentence boundaries by edit category. Overall, 43\% of edits are multi-sentence in nature, confirming that sentence-level simplification overlooks a large fraction of edits. This analysis likely undercounts multi-sentence edits, as anaphora and lexical consistency edits might be applied in a single sentence but require implicit document context.

Each category class occurs in 40-85\% of document pairs (Fig.~\ref{fig:analysis_frequency}). Semantic edits are most common due to the widespread Semantic Deletion category, with all other Semantic categories occurring in 49.6\% of documents.
On average, each annotated document has 15.2 edit operations (6.3 insertions, 8.9 deletions), which are consolidated into 7.8 edit groups (see Figure~\ref{fig:analysis_num_groups} for the full distribution). Non-simplification edits, which correspond to undesirable edits related to formatting, the deletion of noise such as spam edits, or the introduction of extraneous information occur in 64.6\% of document pairs, confirming the noisy nature of Wikipedia-based datasets. In Section~\ref{section:silver_annotation}, we explore an automated cleaning process to remove non-simplification edits.

To understand the diversity of edit categories that occur within each simplified document, we count how many of the four category classes occur jointly in simplified documents. The distribution is plotted in Figure~\ref{fig:analysis_joint_groups}, revealing that a majority of annotated documents contain edits from three or four distinct category classes, confirming that SEW editors combine diverse editing strategies when simplifying EW pages into SEW pages.

We find that individual operations belong to a single group roughly 95\% of the time, meaning that edit group overlap is rare, but find instances of operations belonging to up to 4 groups. Category pairs that overlap most often are (Reordering, Phrasal Edit) and (Reordering, Sentence Splitting).

In summary, the annotated portion of \dataset\ reveals that all four category classes are prevalent on SEW, that at least 43\% of edits require document-level context, and that producing SEW pages often requires combining edits from the full range of edit categories. 

\section{Automatic Edit Identification}
\label{section:identifiers}

We investigate whether edits can be identified automatically, which could automate annotation of the entire \dataset{} dataset -- estimated to require 2,900 hours of manual work -- or facilitate analysis of generative simplification models.

\subsection{Task Definition}
The input to the edit identification task is a document pair's alignment sequence, which is composed of a series of edit operations (Figure~\ref{fig:illustrative_example}); the task is to group (potentially overlapping) edit operations and assign each group to an edit category, matching the format of the annotations. 

Evaluation is performed with four metrics. \textbf{Category F-1} and \textbf{Class F-1} evaluate the predicted categories (19 possible values) and associated higher-level classes (5 possible values) for each edit operation, irrespective of group. We use weighted, multi-label F1 since an edit operation may belong to multiple categories (e.g. for overlapping groups).

The other two metrics consider group assignment and category jointly. \textbf{\%Exact} is the percentage of reference groups for which there is an identical group in the predictions. \textbf{\%Partial} is the percentage of reference groups for which a predicted group of the same category has an operation set overlap of at least 0.5 Jaccard index.

\subsection{Edit Identification Models}

\begin{figure}
    \centering
    \includegraphics[width=.49\textwidth]{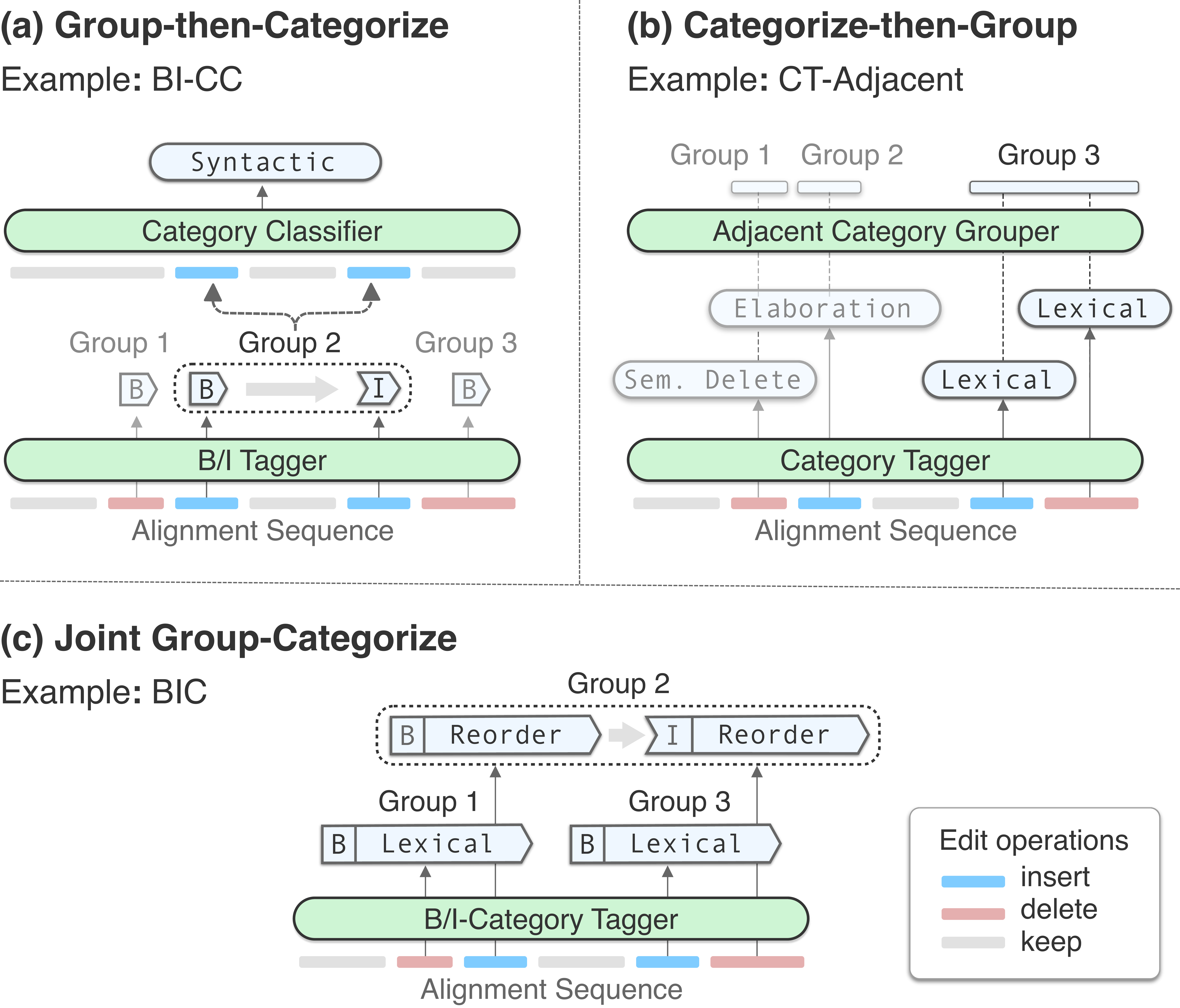}
    \caption{Overview of edit identification models}
    \label{fig:identification_models}
\end{figure}

We implemented three varieties of edit identification models, illustrated in Figure~\ref{fig:identification_models} and described below. Additional details on model architectures are presented in Appendix~\ref{appendix:id_model_specifics}.

The \textit{Group-then-Categorize} approach uses an initial grouper model to propose category-agnostic edit groups, and a second classification model to assign a category to each group (Figure~\ref{fig:identification_models}a). We experiment with three grouper models. The \textit{oracle} grouper uses the groups available in the annotations. The \textit{adjacency} grouper applies the heuristic that adjacent edit operations (with no unchanged text between them) are within the same group. The \textit{BI} grouper is a learned sequence-tagging model that segments edit operations into groups by outputting \textit{B} (Beginning of group) or \textit{I} (Inside of group) for each edit operation. In the next stage, each predicted group is passed to the \textit{Category Classification (CC)} model; the input group is represented as an adjusted alignment sequence in which only the edit operations of the group are included. We refer to the three variants of this two-stage pipeline as \textbf{Oracle-CC}, \textbf{Adjacent-CC}, and \textbf{BI-CC}.

The \textit{Categorize-then-Group} approach first predicts the category of each edit operation and then groups operations based on the predicted categories  (Figure~\ref{fig:identification_models}b). 
For the first stage, we propose \textit{Category Tagger (CT)}, an NER-style sequence tagging model that takes as input a formatted alignment sequence and predicts one or more categories for each edit operation. For the second stage, we explore three grouper models: the \textit{single} grouper performs no grouping, the \textit{adjacent} grouper bundles adjacent edit operations of the same category, and the \textit{rules} grouper applies category-specific rules detailed in Appendix~\ref{appendix:id_model_specifics}. By combining the stages, we obtain \textbf{CT-single}, \textbf{CT-adjacent}, and \textbf{CT-rules}.

In addition to two-stage models, we implemented two \textit{joint} models that simultaneously group and categorize edit operations.
\textbf{BIC} (Fig.~\ref{fig:identification_models}c) is a sequence tagger that combines the label space of the \textbf{BI} and \textbf{C}ategory taggers;
for each edit operation, BIC outputs one or more categories, each paired with a  BI indicator for segmenting groups \textit{within that category}.
This category-specific BI notation supports richer forms of groupings, e.g.,
interleaved groups as illustrated in Figure~\ref{fig:identification_models}c. The \textbf{Seq2seq} model is a fine-tuned sequence-to-sequence model that takes as input an XML-formatted alignment sequence and outputs an expanded XML in which edit categories and groups are identified. 
With all of the above models, we use RoBERTa-large \cite{liu2019roberta} and BART-Large \cite{lewis2020bart} models for NLU and NLG components, respectively. Training details may be found in Appendix~\ref{appendix:id_model_specifics}.

The \textbf{Op Majority} baseline predicts the majority class for each operation type: \textit{Semantic Deletion} for delete operations (54\% of all deletions), and \textit{Lexical} for insert operations (20\% of all insertions).

\subsection{Results}

\begin{table}[]
    \resizebox{0.5\textwidth}{!}{%
    \begin{tabular}{lcccc}
    \textbf{Model Name} & \textbf{Cat F1} & \textbf{Class F1} & \textbf{\%Part} & \textbf{\%Exact}    \\
    \cmidrule(r){1-1} \cmidrule(r){2-3} \cmidrule(r){4-5}
    Op Majority         & 26.1            & 30.3              & -                & -                  \\
    \cmidrule(r){1-1} \cmidrule(r){2-3} \cmidrule(r){4-5}
    Adjacent-CC         & 56.7            & 60.4              & 48.2              & 50.8              \\
    BI-CC               & 64.4            & 67.8              & 56.4             & 60.0               \\
    Oracle-CC           & 78.2            & 81.4              & -                & -                  \\
    \cmidrule(r){1-1} \cmidrule(r){2-3} \cmidrule(r){4-5}
    CT-Single           & 69.7            & \textbf{74.1}      & 27.8             & 27.8               \\
    CT-Adjacent         & 69.7            & \textbf{74.1}      & 58.3             & 60.8               \\
    CT-Rules            & 69.7            & \textbf{74.1}      & 58.4             & 62.1               \\
    \cmidrule(r){1-1} \cmidrule(r){2-3} \cmidrule(r){4-5}
    BIC                 & \textbf{70.6}   & \textbf{74.0}      & \textbf{59.7}    & \textbf{64.7}      \\
    Seq2Seq             & 51.3            & 55.4               & 42.5             & 45.7               \\
    \bottomrule
    \end{tabular}
    }
    \caption{Edit identification results on in-domain test set}
    \label{table:identification_id}
\end{table}

All models were trained on the training set of annotations, and hyperparameters were selected using the validation set. Table~\ref{table:identification_id} summarizes experimental results on the in-domain test set.

Overall, the joint BIC model -- trained to predict grouping and categories together -- achieved the highest performance across the board, showing the benefits of joint over multi-step approaches.
Appendix~\ref{appendix:bic_detail} provides a category-specific breakdown of BIC model performance, revealing that the model excels at identifying edits of common categories (with top-5 F-1 performance coming in the seven most common categories), but struggles with less common categories.

With the Group-then-Categorize models, as grouping quality increases, performance improves as well. When oracle groups are available, the categorization model achieves a 78.2 F-1 score at the category level, indicating that categorizing isolated edits is much less challenging than identifying overlapping edits in entire documents.

The Seq2seq model outperforms the majority baseline, but trails other models, showing that the added flexibility of generative modeling is not beneficial to edit identification in this case.

We report results on the out-of-domain test set in Appendix~\ref{appendix:identification_ood}. We do not observe a consistent performance drop on the unseen Wikipedia categories, giving evidence that most models generalize across categories. In Appendix~\ref{appendix:identification_ood}, we also benchmark the models' computational efficiency and find that BIC performs favorably compared to pipelined approaches and can process 18.9 documents per second on a single GPU, demonstrating another benefit of joint modeling.

\subsection{Dataset Silver Annotation}
\label{section:silver_annotation}
We use the BIC model to automatically annotate all documents in \dataset{}, identifying over one million edits, including more than 90,000 elaborations. Category-specific statistics are in Appendix~\ref{appendix:categories_extras}.

We refine \dataset{} into a \textbf{cleaned} version by automatically reversing edits tagged in the Non-Simplification class. In Section~\ref{section:generators}, we determine whether models trained on the cleaned \dataset{} are less prone to generating unwanted edits, such as ones including extraneous information. 

\section{Text Simplification Baselines}
\label{section:generators}
\begin{figure}
    \centering
    \includegraphics[width=0.45\textwidth]{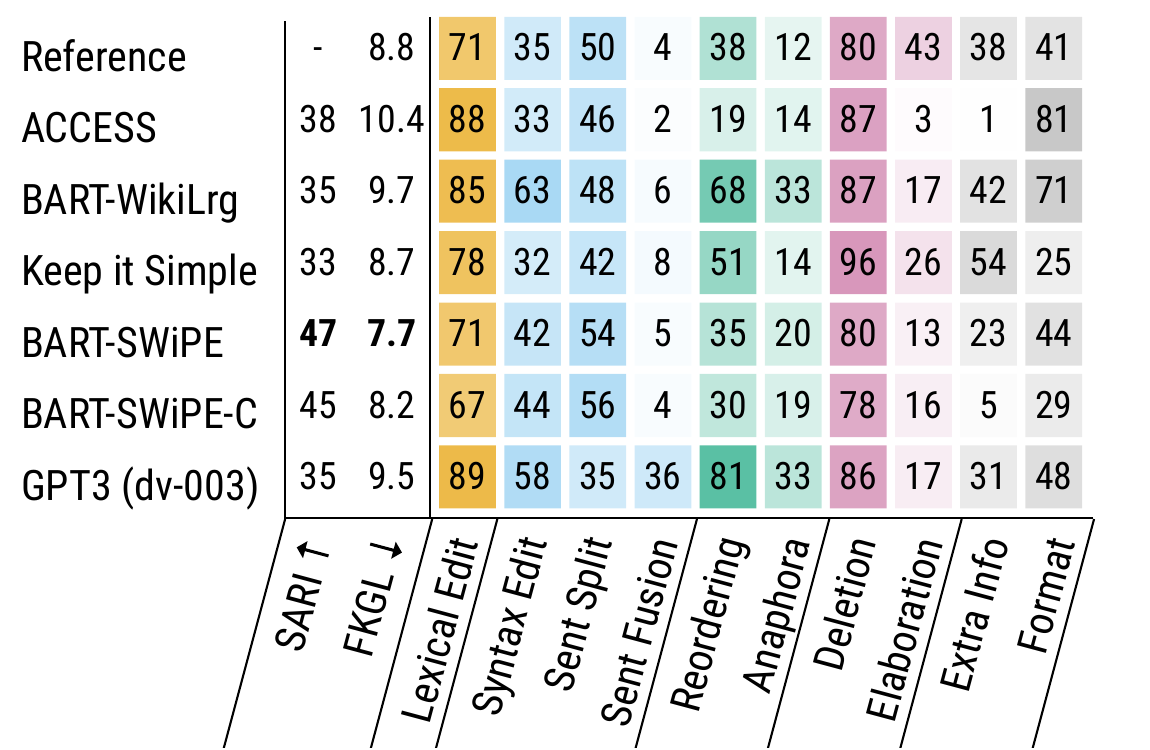}
    \caption{Analysis of generated simplifications: SARI, FKGL, and percentage of identified edit categories.}
    \label{fig:generators_comparison}
\end{figure}

We leverage \dataset{} and its cleaned alternative to fine-tune two BART-large models: \textbf{BART-\dataset{}} and \textbf{BART-\dataset{}-C} and compare them to recent simplification systems.
We experiment with two existing simplification systems: \textbf{ACCESS} \cite{martin2020controllable}, a state-of-the-art controllable sentence-level simplification model trained on Wikilarge \cite{zhang2017sentence}, and \textbf{Keep it Simple} (KIS) \cite{laban2021keep}, an unsupervised paragraph-level model optimized to produce lexical and syntactic edits. We also train \textbf{BART-Wikilarge} a BART-large model trained on Wikilarge to understand the effect of the dataset under a fixed pre-trained model. Finally, we include a prompt-based \textbf{GPT3-davinci-003} using a task prompt that did not specify edit categories to apply. Model details and example outputs are in Appendix~\ref{appendix:genertion_extras}.

We run experiments on the validation set of \dataset{}. For each model, we report the n-gram-based SARI score \cite{xu2016optimizing}, the Flesch-Kincaid Grade Level \cite{Kincaid1975DerivationON}, and the distribution of edit categories identified by BIC (merged into 10 groups). Results are in Figure~\ref{fig:generators_comparison}.

\dataset{}-trained models achieve the highest performance in terms of SARI, confirming a similarity to reference simplifications, and the lowest estimated grade-level scores, validating the model's ability to improve readability.

The ACCESS sentence-level model performs moderately well on the SARI metric, but worst on the grade-level estimation, and makes fewer complex edits such as reorderings or elaborations, confirming that sentence-level models focus on simpler edits, such as lexical and syntactic edits.

All other models attempt a large proportion of all edits, including a large number of edits tagged as extraneous information (i.e., information not in the original document). When simplified by human editors, extraneous information often comes from other documents or background knowledge and is not likely harmful. On the contrary, recent NLG work has shown that model-generated extraneous information is often hallucinated, can be factually incorrect, and is undesirable. Example model outputs in Appendix~\ref{appendix:example_generated_simplifications} show example problematic outputs from the KIS and BART-Wikilarge models which include factual errors, for example confusing centimeters and micrometers, or the length and width of a hair.

The KIS, BART-Wikilarge, BART-\dataset{}, and GPT-3 models all produce a larger proportion of extraneous information edits than elaborations, confirming prior work showing that problematic hallucinations can occur for the simplification task as well \cite{devaraj2022evaluating}. BART-\dataset{}-C is able to produce elaborations while having a reduced rate of extraneous information, giving preliminary evidence that the edit-based dataset cleaning process we adopt can mitigate -- but not solve -- the generation of extraneous information.

Similar to recent work in summarization showing that zero-shot GPT3 can tie or surpass supervised models \cite{goyal2022news, liu2022revisiting}, we observe that GPT3 can generate a wide range of simplification edits and does not mirror priors of the dataset -- such as producing more sentence splits than fusions -- indicating it has potential for use as a general-purpose simplification model. Similar to prior work, GPT3-based candidates score poorly on reference-based metrics.

We note that the analysis is preliminary, and future work should assess the efficacy, factual consistency, and simplicity of generated edits with target readers as done in prior work \cite{laban2021keep} to gain a thorough understanding of model performance.

\section{Discussion \& Future Work}

\textbf{Edit-Based Evaluation of Generators.} In Section~\ref{section:generators}, we compare baseline simplification models based on the types of edits they produce. This analysis is based on automatically identified edits by the BIC model we trained, which likely includes errors. We expect that BIC's errors should affect all of the models' candidates equally, and should not significantly affect overall trends. More manual analysis is required to establish the effectiveness of edits (i.e. whether the applied edits successfully simplify the document), as well as whether edits are factual and reflect the original document's content.

\textbf{Extraneous Information in Simplification.} In Section~\ref{section:silver_annotation}, we create a version of the SWiPE dataset where we remove edits that require extraneous information for a generation. We however choose to release the original dataset which includes those edits, as they could be valuable for future work, for example, approaches that might retrieve relevant documents prior to simplifying or to generate negative samples which can be used to stress-test models \cite{Laban2022NearNegativeDG}.

\textbf{Out-of-Domain Testing.} We created an out-of-domain test set by selecting three Wikipedia categories that would be entirely isolated as a test set, to establish whether models would be capable of generalizing to unseen categories. In Section~\ref{section:identifiers}, we did not observe a meaningful gap in model performance between the in-domain and out-of-domain test sets, indicating that the Wikipedia categories we selected are not dissimilar enough from in-domain categories. Future work could explore other axes to create challenging out-of-domain test sets, for instance, based on page author identity, or publication time.

\textbf{Edit-Based Models.} In Section~\ref{section:generators}, we experiment with models that approach text simplification as a sequence-to-sequence model task and do not explicitly represent the editing process. However, recent progress in text-editing models \cite{malmi2022text} could provide an avenue for better models in text simplification, which could be more efficient computationally and explainable in their generations. It is likely that text-editing models trained for sentence-level simplification \cite{malmi2019encode,agrawal2021non} can be expanded using \dataset{} to generate a wider set of edits that can leverage document-level context.

\textbf{Plan-then-Execute Models.} Prior work in conditional generation tasks such as story generation \cite{martin2018event}, data-to-text generation \cite{puduppully2019data}, or summarization \cite{narayan2021planning} have decomposed the task in two steps, involving first the generation of a high-level plan, followed by an execution step that generates the output conditioned on the desired plan. The \dataset{} resource can enable such research in the field of simplification, as the precise edit-based annotations we collected can serve as a basis for a plan to condition a generation model on. Plan-then-execute models enrich models with an intermediary representation that can be modified by a potential user, enabling customizable simplification applications.

\textbf{Towards Practical Simplification.} Practical implementations of text simplification, such as the news website Newsela \cite{xu2015problems} which simplifies the news to make it accessible to multiple grade-level tiers, require document-level understanding and editing. We hope the \dataset{} dataset and models can play a part in making textual content more accessible, for example by improving access to scientific documents \cite{august2022paper} or news coverage diversity \cite{laban2023designing}.

\section{Conclusion}

We introduce \dataset{}, a large-scale document-level simplification dataset based on Wikipedia. \dataset{} is created by collecting pairs of pages from the English and Simple English Wikipedia and matching their revision histories to build document pairs that align in terms of content presented. We collaborated with Simple Wikipedia editors to annotate 5,000 document pairs in \dataset{}, finding that many complex edits that require document-level context such as elaborations frequently occur in the dataset. We experimented with the automatic identification of edits, finding that even though the task is challenging, some models are able to achieve performance above 0.7 F-1 at edit categorization, making them viable to analyze model-generated simplifications. An analysis of generative simplification models reveals that sentence-level models are limited in the types of edits they propose and that document-scoped models are likely to produce hallucinated content. Finally, a model fine-tuned on a cleaned version of \dataset{} produces less extraneous content while continuing to generate complex edits, pointing towards simplification models that can generate complex yet factually consistent edits.

\section*{Acknowledgments}

We would like to thank the Simple Wikipedia editors and other participants that participated in the data annotation that led to the creation of \dataset{}.

\section{Limitations}

\textbf{\dataset{} focuses on the English language.} Although it is possible that some aspects of the work -- such as the edit categorization -- might transfer to the study of text simplification in other languages, we focus on the English language. As of the writing of this paper, there is no equivalent of Simple English Wikipedia for other languages on Wikipedia, and creating similar resources for other languages would require finding other resources.

\textbf{Difficulty in Retracing Original Editing.} By matching revisions of Wikipedia pages that are factually aligned, and working with SEW editors to annotate the edits, we attempted to match the process used to create the resource. It is however not possible to recruit all 5,000+ SEW editors and for some page pairs the annotations are another editor's best attempt to reconstruct the intended edits by the original editor.

\textbf{Improving Annotation Reproducibility.} The analysis we conduct in Section~\ref{section:categorization} reveals that our annotators achieve moderate agreement on samples repeatedly annotated. More detailed analysis reveals that agreement is generally strong from common edit categories such as Lexical Edits, semantic deletions, or sentence splitting, but is lower for more infrequent categories. Better training of annotators on tail categories could therefore likely improve annotation. We also found that discussion amongst annotators of a sample often led to eventual consensus. Therefore collecting multiple annotations per sample, and allowing for discussion when multiple interpretations occur could help improve annotation quality, but at an increased cost.

\section{Ethical Considerations}

The models and datasets utilized in the project primarily reflect the culture of the English-speaking populace. Gender, age, race, and other socio-economic biases may exist in the dataset, and models trained on these datasets may propagate these biases. Text generation tasks such as simplification have previously been shown to contain these biases.

In our collaboration with Wikipedia Editors to produce the annotations for \dataset{}, we ensured to remunerate the participants fairly (\$25/hour), including for fully or partially completing the onboarding task. Participants could communicate with us to voice concerns, could work at their own pace, and choose to stop working on the project at any time. Finally, we ensured to anonymize the annotations by not including personally identifiable information in any version of the dataset (annotator identity is instead marked as \texttt{annotator1}, \texttt{annotator2}, etc.).

We note that the models we use are imperfect and can make errors. When interpreting our models' outputs, results should be interpreted not in terms of certainty but probability. For example, if one of the simplification models generates edits that introduce non-trivial information, it is possible for this information to be hallucinated and not factually correct. Model outputs should therefore be checked, or a warning that content was machine-generated should be given to the reading audience.

To build the \dataset{} dataset, we relied on several datasets as well as pre-trained language models. We explicitly verified that all datasets and models are publicly released for research purposes and that we have proper permission to reuse and modify the models.

\bibliography{anthology,custom}

\begin{thebibliography}{54}
\expandafter\ifx\csname natexlab\endcsname\relax\def\natexlab#1{#1}\fi

\bibitem[{Agrawal et~al.(2021)Agrawal, Xu, and Carpuat}]{agrawal2021non}
Sweta Agrawal, Weijia Xu, and Marine Carpuat. 2021.
\newblock A non-autoregressive edit-based approach to controllable text
  simplification.
\newblock In \emph{Findings of the Association for Computational Linguistics:
  ACL-IJCNLP 2021}, pages 3757--3769.

\bibitem[{Alu{\'\i}sio et~al.(2008)Alu{\'\i}sio, Specia, Pardo, Maziero,
  Caseli, and Fortes}]{aluisio2008corpus}
Sandra~M Alu{\'\i}sio, Lucia Specia, Thiago~AS Pardo, Erick~G Maziero, Helena~M
  Caseli, and Renata~PM Fortes. 2008.
\newblock A corpus analysis of simple account texts and the proposal of
  simplification strategies: first steps towards text simplification systems.
\newblock In \emph{Proceedings of the 26th annual ACM international conference
  on Design of communication}, pages 15--22.

\bibitem[{Alva-Manchego et~al.(2020)Alva-Manchego, Martin, Bordes, Scarton,
  Sagot, and Specia}]{alva2020asset}
Fernando Alva-Manchego, Louis Martin, Antoine Bordes, Carolina Scarton,
  Beno{\^\i}t Sagot, and Lucia Specia. 2020.
\newblock Asset: A dataset for tuning and evaluation of sentence simplification
  models with multiple rewriting transformations.
\newblock In \emph{Proceedings of the 58th Annual Meeting of the Association
  for Computational Linguistics}, pages 4668--4679.

\bibitem[{August et~al.(2022)August, Wang, Bragg, Hearst, Head, and
  Lo}]{august2022paper}
Tal August, Lucy~Lu Wang, Jonathan Bragg, Marti~A Hearst, Andrew Head, and Kyle
  Lo. 2022.
\newblock Paper plain: Making medical research papers approachable to
  healthcare consumers with natural language processing.
\newblock \emph{arXiv preprint arXiv:2203.00130}.

\bibitem[{Boyd(2018)}]{boyd2018using}
Adriane Boyd. 2018.
\newblock Using wikipedia edits in low resource grammatical error correction.
\newblock In \emph{Proceedings of the 2018 EMNLP Workshop W-NUT: The 4th
  Workshop on Noisy User-generated Text}, pages 79--84.

\bibitem[{Cardon et~al.(2022)Cardon, Bibal, Souza~Wilkens, Alfter, Norr{\'e},
  M{\"u}ller, Watrin, and Fran{\c{c}}ois}]{cardon2022linguistic}
R{\'e}mi Cardon, Adrien Bibal, Rodrigo Souza~Wilkens, David Alfter, Magali
  Norr{\'e}, Adeline M{\"u}ller, Patrick Watrin, and Thomas Fran{\c{c}}ois.
  2022.
\newblock Linguistic corpus annotation for automatic text simplification
  evaluation.
\newblock In \emph{EMNLP 2022}.

\bibitem[{Chin et~al.(2010)Chin, Street, Srinivasan, and
  Eichmann}]{chin2010detecting}
Si-Chi Chin, W~Nick Street, Padmini Srinivasan, and David Eichmann. 2010.
\newblock Detecting wikipedia vandalism with active learning and statistical
  language models.
\newblock In \emph{Proceedings of the 4th workshop on Information credibility},
  pages 3--10.

\bibitem[{Coster and Kauchak(2011)}]{coster2011simple}
William Coster and David Kauchak. 2011.
\newblock Simple english wikipedia: a new text simplification task.
\newblock In \emph{Proceedings of the 49th Annual Meeting of the Association
  for Computational Linguistics: Human Language Technologies}, pages 665--669.

\bibitem[{Devaraj et~al.(2021)Devaraj, Marshall, Wallace, and
  Li}]{devaraj2021paragraph}
Ashwin Devaraj, Iain Marshall, Byron~C Wallace, and Junyi~Jessy Li. 2021.
\newblock Paragraph-level simplification of medical texts.
\newblock In \emph{Proceedings of the 2021 Conference of the North American
  Chapter of the Association for Computational Linguistics: Human Language
  Technologies}, pages 4972--4984.

\bibitem[{Devaraj et~al.(2022)Devaraj, Sheffield, Wallace, and
  Li}]{devaraj2022evaluating}
Ashwin Devaraj, William Sheffield, Byron~C Wallace, and Junyi~Jessy Li. 2022.
\newblock Evaluating factuality in text simplification.
\newblock In \emph{Proceedings of the 60th Annual Meeting of the Association
  for Computational Linguistics (Volume 1: Long Papers)}, pages 7331--7345.

\bibitem[{Dutrey et~al.(2010)Dutrey, Bouamor, Bernhard, and
  Max}]{dutrey2010local}
Camille Dutrey, Houda Bouamor, Delphine Bernhard, and Aur{\'e}lien Max. 2010.
\newblock Local modifications and paraphrases in wikipedia's revision history.
\newblock \emph{Procesamiento del lenguaje natural}, 46:51--58.

\bibitem[{Dwivedi-Yu et~al.(2022)Dwivedi-Yu, Schick, Jiang, Lomeli, Lewis,
  Izacard, Grave, Riedel, and Petroni}]{dwivedi2022editeval}
Jane Dwivedi-Yu, Timo Schick, Zhengbao Jiang, Maria Lomeli, Patrick Lewis,
  Gautier Izacard, Edouard Grave, Sebastian Riedel, and Fabio Petroni. 2022.
\newblock Editeval: An instruction-based benchmark for text improvements.
\newblock \emph{arXiv preprint arXiv:2209.13331}.

\bibitem[{Glava{\v{s}} and {\v{S}}tajner(2015)}]{glavavs2015simplifying}
Goran Glava{\v{s}} and Sanja {\v{S}}tajner. 2015.
\newblock Simplifying lexical simplification: Do we need simplified corpora?
\newblock In \emph{Proceedings of the 53rd Annual Meeting of the Association
  for Computational Linguistics and the 7th International Joint Conference on
  Natural Language Processing (Volume 2: Short Papers)}, pages 63--68.

\bibitem[{Goyal et~al.(2022)Goyal, Li, and Durrett}]{goyal2022news}
Tanya Goyal, Junyi~Jessy Li, and Greg Durrett. 2022.
\newblock News summarization and evaluation in the era of gpt-3.
\newblock \emph{arXiv preprint arXiv:2209.12356}.

\bibitem[{Heindorf et~al.(2015)Heindorf, Potthast, Stein, and
  Engels}]{heindorf2015towards}
Stefan Heindorf, Martin Potthast, Benno Stein, and Gregor Engels. 2015.
\newblock Towards vandalism detection in knowledge bases: Corpus construction
  and analysis.
\newblock In \emph{Proceedings of the 38th International ACM SIGIR Conference
  on Research and Development in Information Retrieval}, pages 831--834.

\bibitem[{Honnibal et~al.(2020)Honnibal, Montani, Van~Landeghem, and
  Boyd}]{spacy2020}
Matthew Honnibal, Ines Montani, Sofie Van~Landeghem, and Adriane Boyd. 2020.
\newblock \href {https://doi.org/10.5281/zenodo.1212303} {{spaCy:
  Industrial-strength Natural Language Processing in Python}}.

\bibitem[{Jiang et~al.(2020)Jiang, Maddela, Lan, Zhong, and
  Xu}]{jiang2020neural}
Chao Jiang, Mounica Maddela, Wuwei Lan, Yang Zhong, and Wei Xu. 2020.
\newblock Neural crf model for sentence alignment in text simplification.
\newblock In \emph{Proceedings of the 58th Annual Meeting of the Association
  for Computational Linguistics}, pages 7943--7960.

\bibitem[{Jiang et~al.(2022)Jiang, Xu, and Stevens}]{jiang2022arxivedits}
Chao Jiang, Wei Xu, and Samuel Stevens. 2022.
\newblock arxivedits: Understanding the human revision process in scientific
  writing.
\newblock \emph{arXiv preprint arXiv:2210.15067}.

\bibitem[{Kincaid et~al.(1975)Kincaid, Fishburne~Jr., Rogers, and
  Chissom}]{Kincaid1975DerivationON}
J.~Peter Kincaid, Robert~P. Fishburne~Jr., Richard~L. Rogers, and Brad~S.
  Chissom. 1975.
\newblock Derivation of new readability formulas (automated readability index,
  fog count and flesch reading ease formula) for navy enlisted personnel.
\newblock Technical report, Naval Technical Training Command Millington TN
  Research Branch.

\bibitem[{Kry{\'s}ci{\'n}ski et~al.(2020)Kry{\'s}ci{\'n}ski, McCann, Xiong, and
  Socher}]{kryscinski2020evaluating}
Wojciech Kry{\'s}ci{\'n}ski, Bryan McCann, Caiming Xiong, and Richard Socher.
  2020.
\newblock Evaluating the factual consistency of abstractive text summarization.
\newblock In \emph{Proceedings of the 2020 Conference on Empirical Methods in
  Natural Language Processing (EMNLP)}, pages 9332--9346.

\bibitem[{Laban et~al.(2021)Laban, Schnabel, Bennett, and
  Hearst}]{laban2021keep}
Philippe Laban, Tobias Schnabel, Paul Bennett, and Marti~A Hearst. 2021.
\newblock Keep it simple: Unsupervised simplification of multi-paragraph text.
\newblock In \emph{Proceedings of the 59th Annual Meeting of the Association
  for Computational Linguistics and the 11th International Joint Conference on
  Natural Language Processing (Volume 1: Long Papers)}, pages 6365--6378.

\bibitem[{Laban et~al.(2022{\natexlab{a}})Laban, Schnabel, Bennett, and
  Hearst}]{laban2022summac}
Philippe Laban, Tobias Schnabel, Paul~N Bennett, and Marti~A Hearst.
  2022{\natexlab{a}}.
\newblock Summac: Re-visiting nli-based models for inconsistency detection in
  summarization.
\newblock \emph{Transactions of the Association for Computational Linguistics},
  10:163--177.

\bibitem[{Laban et~al.(2022{\natexlab{b}})Laban, Wu, Liu, and
  Xiong}]{Laban2022NearNegativeDG}
Philippe Laban, Chien-Sheng Wu, Wenhao Liu, and Caiming Xiong.
  2022{\natexlab{b}}.
\newblock Near-negative distinction: Giving a second life to human evaluation
  datasets.
\newblock In \emph{Conference on Empirical Methods in Natural Language
  Processing}.

\bibitem[{Laban et~al.(2023)Laban, Wu, Murakhovs'~Ka, Chen, and
  Xiong}]{laban2023designing}
Philippe Laban, Chien-Sheng Wu, Lidiya Murakhovs'~Ka, Xiang'Anthony' Chen, and
  Caiming Xiong. 2023.
\newblock Designing and evaluating interfaces that highlight news coverage
  diversity using discord questions.
\newblock In \emph{Proceedings of the 2023 CHI Conference on Human Factors in
  Computing Systems}, pages 1--21.

\bibitem[{Levenshtein(1966)}]{Levenshtein1966BinaryCC}
Vladimir~I. Levenshtein. 1966.
\newblock Binary codes capable of correcting deletions, insertions and
  reversals.
\newblock \emph{Soviet physics. Doklady}, 10:707--710.

\bibitem[{Lewis et~al.(2020)Lewis, Liu, Goyal, Ghazvininejad, Mohamed, Levy,
  Stoyanov, and Zettlemoyer}]{lewis2020bart}
Mike Lewis, Yinhan Liu, Naman Goyal, Marjan Ghazvininejad, Abdelrahman Mohamed,
  Omer Levy, Veselin Stoyanov, and Luke Zettlemoyer. 2020.
\newblock Bart: Denoising sequence-to-sequence pre-training for natural
  language generation, translation, and comprehension.
\newblock In \emph{Proceedings of the 58th Annual Meeting of the Association
  for Computational Linguistics}, pages 7871--7880.

\bibitem[{Liu et~al.(2019)Liu, Ott, Goyal, Du, Joshi, Chen, Levy, Lewis,
  Zettlemoyer, and Stoyanov}]{liu2019roberta}
Yinhan Liu, Myle Ott, Naman Goyal, Jingfei Du, Mandar Joshi, Danqi Chen, Omer
  Levy, Mike Lewis, Luke Zettlemoyer, and Veselin Stoyanov. 2019.
\newblock Roberta: A robustly optimized bert pretraining approach.
\newblock \emph{arXiv preprint arXiv:1907.11692}.

\bibitem[{Liu et~al.(2022)Liu, Fabbri, Liu, Zhao, Nan, Han, Han, Joty, Wu,
  Xiong et~al.}]{liu2022revisiting}
Yixin Liu, Alexander~R Fabbri, Pengfei Liu, Yilun Zhao, Linyong Nan, Ruilin
  Han, Simeng Han, Shafiq Joty, Chien-Sheng Wu, Caiming Xiong, et~al. 2022.
\newblock Revisiting the gold standard: Grounding summarization evaluation with
  robust human evaluation.
\newblock \emph{arXiv preprint arXiv:2212.07981}.

\bibitem[{Malmi et~al.(2022)Malmi, Dong, Mallinson, Chuklin, Adamek, Mirylenka,
  Stahlberg, Krause, Kumar, and Severyn}]{malmi2022text}
Eric Malmi, Yue Dong, Jonathan Mallinson, Aleksandr Chuklin, Jakub Adamek,
  Daniil Mirylenka, Felix Stahlberg, Sebastian Krause, Shankar Kumar, and
  Aliaksei Severyn. 2022.
\newblock Text generation with text-editing models.
\newblock \emph{arXiv preprint arXiv:2206.07043}.

\bibitem[{Malmi et~al.(2019)Malmi, Krause, Rothe, Mirylenka, and
  Severyn}]{malmi2019encode}
Eric Malmi, Sebastian Krause, Sascha Rothe, Daniil Mirylenka, and Aliaksei
  Severyn. 2019.
\newblock Encode, tag, realize: High-precision text editing.
\newblock In \emph{Proceedings of the 2019 Conference on Empirical Methods in
  Natural Language Processing and the 9th International Joint Conference on
  Natural Language Processing (EMNLP-IJCNLP)}, pages 5054--5065.

\bibitem[{Martin et~al.(2018)Martin, Ammanabrolu, Wang, Hancock, Singh,
  Harrison, and Riedl}]{martin2018event}
Lara Martin, Prithviraj Ammanabrolu, Xinyu Wang, William Hancock, Shruti Singh,
  Brent Harrison, and Mark Riedl. 2018.
\newblock Event representations for automated story generation with deep neural
  nets.
\newblock In \emph{Proceedings of the AAAI Conference on Artificial
  Intelligence}, volume~32.

\bibitem[{Martin et~al.(2020)Martin, De~La~Clergerie, Sagot, and
  Bordes}]{martin2020controllable}
Louis Martin, {\'E}ric~Villemonte De~La~Clergerie, Beno{\^\i}t Sagot, and
  Antoine Bordes. 2020.
\newblock Controllable sentence simplification.
\newblock In \emph{Proceedings of the 12th Language Resources and Evaluation
  Conference}, pages 4689--4698.

\bibitem[{Max and Wisniewski(2010)}]{max2010mining}
Aur{\'e}lien Max and Guillaume Wisniewski. 2010.
\newblock Mining naturally-occurring corrections and paraphrases from
  wikipedia’s revision history.
\newblock In \emph{Proceedings of the Seventh International Conference on
  Language Resources and Evaluation (LREC'10)}.

\bibitem[{Maynez et~al.(2020)Maynez, Narayan, Bohnet, and
  McDonald}]{maynez2020faithfulness}
Joshua Maynez, Shashi Narayan, Bernd Bohnet, and Ryan McDonald. 2020.
\newblock On faithfulness and factuality in abstractive summarization.
\newblock In \emph{Proceedings of the 58th Annual Meeting of the Association
  for Computational Linguistics}, pages 1906--1919.

\bibitem[{Narayan and Gardent(2015)}]{Narayan2015UnsupervisedSS}
Shashi Narayan and Claire Gardent. 2015.
\newblock Unsupervised sentence simplification using deep semantics.
\newblock In \emph{International Conference on Natural Language Generation}.

\bibitem[{Narayan et~al.(2021)Narayan, Zhao, Maynez, Sim{\~o}es, Nikolaev, and
  McDonald}]{narayan2021planning}
Shashi Narayan, Yao Zhao, Joshua Maynez, Gon{\c{c}}alo Sim{\~o}es, Vitaly
  Nikolaev, and Ryan McDonald. 2021.
\newblock Planning with learned entity prompts for abstractive summarization.
\newblock \emph{Transactions of the Association for Computational Linguistics},
  9:1475--1492.

\bibitem[{Nelken and Yamangil(2008)}]{nelken2008mining}
Rani Nelken and Elif Yamangil. 2008.
\newblock Mining wikipedia’s article revision history for training
  computational linguistics algorithms.
\newblock In \emph{Proceedings of the AAAI Workshop on Wikipedia and Artificial
  Intelligence: An Evolving Synergy}, pages 31--36.

\bibitem[{Puduppully et~al.(2019)Puduppully, Dong, and
  Lapata}]{puduppully2019data}
Ratish Puduppully, Li~Dong, and Mirella Lapata. 2019.
\newblock Data-to-text generation with content selection and planning.
\newblock In \emph{Proceedings of the AAAI conference on artificial
  intelligence}, volume~33, pages 6908--6915.

\bibitem[{Rajpurkar et~al.(2016)Rajpurkar, Zhang, Lopyrev, and
  Liang}]{Rajpurkar2016SQuAD1Q}
Pranav Rajpurkar, Jian Zhang, Konstantin Lopyrev, and Percy Liang. 2016.
\newblock Squad: 100,000+ questions for machine comprehension of text.
\newblock In \emph{Conference on Empirical Methods in Natural Language
  Processing}.

\bibitem[{Scarton and Specia(2018)}]{scarton2018learning}
Carolina Scarton and Lucia Specia. 2018.
\newblock Learning simplifications for specific target audiences.
\newblock In \emph{Proceedings of the 56th Annual Meeting of the Association
  for Computational Linguistics (Volume 2: Short Papers)}, pages 712--718.

\bibitem[{Schick et~al.(2022)Schick, Dwivedi-Yu, Jiang, Petroni, Lewis,
  Izacard, You, Nalmpantis, Grave, and Riedel}]{schick2022peer}
Timo Schick, Jane Dwivedi-Yu, Zhengbao Jiang, Fabio Petroni, Patrick Lewis,
  Gautier Izacard, Qingfei You, Christoforos Nalmpantis, Edouard Grave, and
  Sebastian Riedel. 2022.
\newblock Peer: A collaborative language model.
\newblock \emph{arXiv preprint arXiv:2208.11663}.

\bibitem[{Schuster et~al.(2021)Schuster, Fisch, and Barzilay}]{schuster2021get}
Tal Schuster, Adam Fisch, and Regina Barzilay. 2021.
\newblock Get your vitamin c! robust fact verification with contrastive
  evidence.
\newblock In \emph{Proceedings of the 2021 Conference of the North American
  Chapter of the Association for Computational Linguistics: Human Language
  Technologies}, pages 624--643.

\bibitem[{Siddharthan(2014)}]{siddharthan2014survey}
Advaith Siddharthan. 2014.
\newblock A survey of research on text simplification.
\newblock \emph{ITL-International Journal of Applied Linguistics},
  165(2):259--298.

\bibitem[{Srikanth and Li(2020)}]{srikanth2020elaborative}
Neha Srikanth and Junyi~Jessy Li. 2020.
\newblock Elaborative simplification: Content addition and explanation
  generation in text simplification.
\newblock \emph{arXiv preprint arXiv:2010.10035}.

\bibitem[{Srikanth and Li(2021)}]{srikanth2021elaborative}
Neha Srikanth and Junyi~Jessy Li. 2021.
\newblock Elaborative simplification: Content addition and explanation
  generation in text simplification.
\newblock In \emph{Findings of the Association for Computational Linguistics:
  ACL-IJCNLP 2021}, pages 5123--5137.

\bibitem[{{\v{S}}tajner et~al.(2018){\v{S}}tajner, Franco-Salvador, Rosso, and
  Ponzetto}]{vstajner2018cats}
Sanja {\v{S}}tajner, Marc Franco-Salvador, Paolo Rosso, and Simone~Paolo
  Ponzetto. 2018.
\newblock Cats: A tool for customized alignment of text simplification corpora.
\newblock In \emph{Proceedings of the Eleventh International Conference on
  Language Resources and Evaluation (LREC 2018)}.

\bibitem[{Sulem et~al.(2018)Sulem, Abend, and Rappoport}]{sulem2018bleu}
Elior Sulem, Omri Abend, and Ari Rappoport. 2018.
\newblock Bleu is not suitable for the evaluation of text simplification.
\newblock In \emph{Proceedings of the 2018 Conference on Empirical Methods in
  Natural Language Processing}, pages 738--744.

\bibitem[{Sun et~al.(2021)Sun, Jin, and Wan}]{sun2021document}
Renliang Sun, Hanqi Jin, and Xiaojun Wan. 2021.
\newblock Document-level text simplification: Dataset, criteria and baseline.
\newblock In \emph{Proceedings of the 2021 Conference on Empirical Methods in
  Natural Language Processing}, pages 7997--8013.

\bibitem[{V{\'a}squez-Rodr{\'\i}guez et~al.(2021)V{\'a}squez-Rodr{\'\i}guez,
  Shardlow, and Ananiadou}]{vasquez2021role}
Laura V{\'a}squez-Rodr{\'\i}guez, Matthew Shardlow, and Sophia Ananiadou. 2021.
\newblock The role of text simplification operations in evaluation.

\bibitem[{Xu et~al.(2015)Xu, Callison-Burch, and Napoles}]{xu2015problems}
Wei Xu, Chris Callison-Burch, and Courtney Napoles. 2015.
\newblock Problems in current text simplification research: New data can help.
\newblock \emph{Transactions of the Association for Computational Linguistics},
  3:283--297.

\bibitem[{Xu et~al.(2016)Xu, Napoles, Pavlick, Chen, and
  Callison-Burch}]{xu2016optimizing}
Wei Xu, Courtney Napoles, Ellie Pavlick, Quanze Chen, and Chris Callison-Burch.
  2016.
\newblock Optimizing statistical machine translation for text simplification.
\newblock \emph{Transactions of the Association for Computational Linguistics},
  4:401--415.

\bibitem[{Zhang and Lapata(2017)}]{zhang2017sentence}
Xingxing Zhang and Mirella Lapata. 2017.
\newblock Sentence simplification with deep reinforcement learning.
\newblock In \emph{Proceedings of the 2017 Conference on Empirical Methods in
  Natural Language Processing}, pages 584--594.

\bibitem[{Zhong et~al.(2020)Zhong, Jiang, Xu, and Li}]{zhong2020discourse}
Yang Zhong, Chao Jiang, Wei Xu, and Junyi~Jessy Li. 2020.
\newblock Discourse level factors for sentence deletion in text simplification.
\newblock In \emph{Proceedings of the AAAI Conference on Artificial
  Intelligence}, volume~34, pages 9709--9716.

\bibitem[{Zhu et~al.(2010)Zhu, Bernhard, and Gurevych}]{zhu2010monolingual}
Zhemin Zhu, Delphine Bernhard, and Iryna Gurevych. 2010.
\newblock A monolingual tree-based translation model for sentence
  simplification.
\newblock In \emph{Proceedings of the 23rd International Conference on
  Computational Linguistics (Coling 2010)}, pages 1353--1361.

\end{thebibliography}
\bibliographystyle{acl_natbib}

\appendix
\label{sec:appendix}
\section{Revision Matching}
\label{appendix:revision_matching}

Given a single revision of a SEW page, the task objective is to identify revisions of the matching EW page that could have been used as a starting point by a Wikipedia editor.

To gain a better understanding of the task at hand, we manually annotated a subset of 2,000 revision pairs from the created dataset. Prior work for sentence-level alignment has shown a relationship between content alignment and shallow string alignment (such as Levenshtein distance). To determine whether string-alignment methods are adequate for document-level alignment, we annotated samples across the entire range of string-alignment similarities, annotating 200 document pairs in each 0.1 range of Levenshtein ratio between [0,1.0].

Revision pairs were annotated by the authors of the paper with binary \texttt{Aligned}/\texttt{Unaligned} labels. A document pair was assigned the \texttt{Aligned} label if all the information in the SEW document was mentioned in the EW document, or if any new information can be seen as a useful addition for the purpose of simplifying information present both in the SEW and EW pages. The most common reason for a document pair to be marked as \texttt{Unaligned} is when the SEW document contains additional sentences or paragraphs that provide information that does not directly assist the information on the EW page.

The annotated data were randomly split into training, validation, and testing splits (1400-300-300 examples). We experimented with a diverse set of zero-shot and supervised methods for the task of page-pair alignment prediction, which we briefly introduce below. For models that predict real-valued scores, we selected a threshold based on the best validation performance.

\noindent\textbf{Baselines.} \texttt{Majority} always predicts the majority class (Aligned), and \texttt{$\Delta$Publish} produces a score based on the difference in publication time of the two revisions.

\noindent\textbf{String-Alignment.} \texttt{Levenshtein Ratio} is the negated normalized Levenshtein distance, and \texttt{Partial Levenshtein Ratio} finds the longest common subsequence (LCS) between the two documents, and computes the LCS's Levenshtein Ratio, allowing penalty-free deletion/insertions at the extrema of either document.

\noindent\textbf{Entity-based.} \texttt{Entity Overlap} uses spaCy's NER model \cite{spacy2020} to extract named entities from both revisions and computes the Jaccard index between the entity sets as a score, with the assumption that newly introduced entities can be a signal of new and unaligned information.

\noindent\textbf{NLI-based.} NLI models such as the SummaC model \cite{laban2022summac} have been successfully adapted to semantic similarity tasks, such as factual inconsistency detection in summarization. We include \texttt{SummaC\textsubscript{Doc}} in our experiments.

\noindent\textbf{Supervised.} We finetune a \texttt{RoBERTa-Large} on the 1,400 training samples, and select the final model based on the checkpoint that achieved the highest F1 score of 82.8 on the validation set.

\begin{table}[]
    \resizebox{0.5\textwidth}{!}{%
    \begin{tabular}{lcccccc}
     & \multicolumn{3}{c}{\textbf{Validation}} & \multicolumn{3}{c}{\textbf{Test}} \\
    \cmidrule(r){2-4} \cmidrule(r){5-7}
    \textbf{Model Name} & \textbf{P} & \textbf{R} & \textbf{F1} & \textbf{P} & \textbf{R} & \textbf{F1} \\
    \cmidrule(r){1-1} \cmidrule(r){2-4} \cmidrule(r){5-7}
    Majority & 59.0 & 100.0 & 74.2 & 62.6 & 100.0 & 77.0 \\
    $\Delta$Publish & 60.1 & 97.7 & 74.2 & 63.3 & 96.8 & 76.6 \\
    \cmidrule(r){1-1} \cmidrule(r){2-4} \cmidrule(r){5-7}
    Lev. Ratio & 63.1 & 100.0 & 76.3 & 65.9 & 96.3 & 78.3 \\
    Partial Lev. R & 64.9 & 97.2 & 77.8 & 66.7 & 94.2 & 78.1 \\
    \cmidrule(r){1-1} \cmidrule(r){2-4} \cmidrule(r){5-7}
    Ent. Overlap & 79.8 & 82.5 & 81.1 & 75.9 & 75.1 & 75.5 \\
    \cmidrule(r){1-1} \cmidrule(r){2-4} \cmidrule(r){5-7}
    SummaC\textsubscript{Doc} & 77.0 & 92.7 & 84.1 & 77.9 & 91.5 & \textbf{84.2} \\
    \cmidrule(r){1-1} \cmidrule(r){2-4} \cmidrule(r){5-7}
    Supervised & 88.9 & 81.4 & \textbf{85.0} & 83.9 & 85.2 & \textbf{84.5} \\
    \bottomrule
    \end{tabular}
    }
    \caption{Performance of models on the page-pair alignment task. Top-to-bottom: baselines, string alignment, NER, NLI, and supervised models. Precision, recall, and F-1 reported on validation and test sets.}
    \label{table:alignment_results}
\end{table}

Table~\ref{table:alignment_results} summarizes results. The $\Delta$Publish and Levenshtein-based methods only narrowly outperform the majority class baseline in terms of F1 performance, confirming recent findings on the limitations of string-based similarity measures \cite{jiang2020neural}. Entity Overlap performs moderately strongly on the validation set but fails to generalize on the test set. Finally, the NLI-based SummaC model and the supervised model both largely outperform other models, and both achieve test F-1 scores of around 84.

We select the SummaC model \cite{laban2022summac} for the dataset creation process, as it achieves similar performance to the supervised model in terms of F1, but with a higher recall (and lower precision). We favor recall for this application, as it will lead to a potentially larger dataset. We note that this choice might come at the cost of some of the samples in the dataset not being high-quality matches.

\section{\dataset{} Edit Definitions}
\label{appendix:category_definitions}
Below is a reproduction of the definitions provided during the onboarding of annotators. 

\subsection{Introduction.}
Edits can be attributed to one of four high-level goals:
\begin{itemize}
    \item \textbf{Lexical} edits are focused on simplifying word units, replacing rare/technical terms – a single word or a phrase – with simpler/more familiar terms.
    \item \textbf{Syntactic} edits are focused on simplifying sentence units, simplifying the structure of a sentence, for example shortening sentences, or reordering clauses within a sentence.
    \item \textbf{Discourse} edits deal with multi-sentence-level understanding, for instance by making connections between sentences more explicit, or reordering content so that required information appears before advanced information.
    \item \textbf{Semantic} edits deal with the addition or removal of information to improve readability at the document level, for example through the deletion of information that is not needed for a preliminary understanding of a document, or elaborations that introduce needed background or practical examples to help a broader audience understand the document.
\end{itemize}

Any edit that does not fit any of the primary simplification goals is categorized as a Non-simplification. Other edits are typically artifacts of the dataset, for example, a fact correction in Wikipedia revisions, or format cleaning (change of spelling or capitalization).

We now give a definition of each edit. Annotators were additionally provided a canonical example of each category, which we omit in the paper, but will include upon publication on an open-source repository.

\subsection{Lexical Edits}

\begin{itemize}
    \item \textbf{Lexical - Entity}. Any edit that specifically targets the simplification of an entity (person, organization, location) for example the removal of a person’s middle name or the replacement of a scientific name with a common name.
    \item \textbf{Lexical}. Any edit that replaces a complex or technical word or phrase with a more common/simple/accessible word or phrase. If the target phrase is a named entity, then the edit should be labeled with the more specific \textbf{Lexical - Entity}.
\end{itemize}

\subsection{Syntactic Edits}

\begin{itemize}
    \item \textbf{Sentence Split.} An edit that leads to a single sentence being divided into two or more shorter sentences. In order for the split to be fluent, words are typically removed and inserted at the sentence boundary. If non-connector content is added, then it is not only a sentence split.
    \item \textbf{Sentence Fusion.} An edit that leads to several (two or more) sentences being merged into a single (potentially longer) sentence. Content is typically removed from original sentences to join the sentences fluently.
    \item \textbf{Syntactic Deletion.} An edit that deletes words in a sentence with the primary objective of compressing the sentence but does not remove information. If information is removed, see \textbf{Semantic - Deletion}.
    \item \textbf{Syntactic Generic.} An edit that modifies the syntax of the sentence, for example through re-ordering of clauses or changing verb tense.
\end{itemize}

\subsection{Discourse Edits}

\begin{itemize}
    \item \textbf{Reordering.} An edit (or typically several edits) that re-orders content to improve narrative flow, for example moving up background content to ease comprehension. The re-ordering can happen within a single sentence, or across multiple sentences.
    \item \textbf{Anaphora Resolution.} An edit that replaces the repeated or implicit mention of an entity – typically a pronoun – with a resolved mention of the entity (i.e., that doesn’t require prior context).
    \item \textbf{Anaphora Insertion.} An edit that replaces an explicit mention of an entity with an indirect mention, such as a pronoun. The pronoun is typically a short common, which can reduce sentence complexity by decreasing length and word complexity. Note: this is the inverse of the \textbf{Anaphora Resolution} edit.
\end{itemize}

\subsection{Semantic Edits}

\begin{itemize}
    \item \textbf{Specific-to-General.} An edit that substitutes or removes low-level detail in exchange for a higher-level description (like replacing a city with its country). The detail deletion typically is judged as not essential and can be replaced by the higher-level portion. There must be a high-level content addition, otherwise, if it is only deletion, it is likely a \textbf{Semantic - Deletion}.
    \item \textbf{Elaboration - Background.} An edit that inserts content – a phrase or a full sentence – adding pre-requisite information for related content in the document. Typically, the background is inserted before the content it supplements.
    \item \textbf{Elaboration - Example.} An edit that inserts a concrete example of an abstract concept or phenomenon described in the document. Typically, the example is inserted after the content it concretizes.
    \item \textbf{Elaboration - Generic.} Any edit that adds information but cannot be categorized as a ``Background'' or ``Example'' elaboration. The insertion can be a phrase or a full sentence.
    \item \textbf{Semantic - Deletion.} An edit that removes content from the original document, typically because it is not essential to a simple comprehension of the document. The deletion can remove a part of a sentence or an entire sentence. Note that there can be many deletions within a single document, particularly when the original document is lengthy.
\end{itemize}

\subsection{Non-Simplification Edits}

\begin{itemize}
    \item \textbf{Format.} An edit that modifies solely the formatting of the document, including punctuation, capitalization, spelling (for example UK to US spelling), or entity format (such as a date).
    \item \textbf{Noise Deletion.} An edit that fixes noisy content in the original document, such as a trailing partial sentence, or Wikipedia-specific formatting and jargon.
    \item \textbf{Fact Correction.} An edit that corrects a specific fact in the original document, most often updating the recency of the fact.
    \item \textbf{Extraneous Information.} Any edit that introduces facts that are not meant to simplify or add context to the information already present. Typically adds related but secondary information that is not needed in the simplified text. The insertion could be within a sentence or an entire sentence.
    \item \textbf{NonSim - General.} Any other edit that does not contribute to (Lexical, Syntactic, Discourse, Semantic) simplification, but does not fit in any other category.
\end{itemize}

\section{Agreement Level \& Silver Statistics}
\label{appendix:categories_extras}

\begin{table}[]
    \resizebox{0.5\textwidth}{!}{%
    \begin{tabular}{lccccc}
    & \multicolumn{2}{c}{\textbf{Manual}} & \multicolumn{2}{c}{\textbf{Silver}} & \\
    \cmidrule(lr){2-3}  \cmidrule(lr){4-5}
    \textbf{Edit Category} & \textbf{N} & \textbf{\%$\exists$} & \textbf{N} &  \textbf{\%$\exists$} & $\kappa$ \\
    \cmidrule(r){1-1} \cmidrule(lr){2-3}  \cmidrule(lr){4-5} \cmidrule(l){6-6}
    {\Large \color{colorlex} $\bullet$} Lexical Edit & 6789 & 61.7 & 246k & 62.0 & 0.62 \\
    {\Large \color{colorlex} $\bullet$} Entity Edit & 359 & 6.4 & 9553 & 5.7 & 0.36  \\
    \cmidrule(r){1-1} \cmidrule(lr){2-3}  \cmidrule(lr){4-5} \cmidrule(l){6-6}
    {\Large \color{colorsyn} $\bullet$} Sentence Split & 3010 & 43.8 & 93k & 41.1 & 0.83 \\
    {\Large \color{colorsyn} $\bullet$} Sentence Fusion & 334 & 6.0 & 8141 & 4.6 & 0.34 \\
    {\Large \color{colorsyn} $\bullet$} Syntactic Deletion & 1889 & 28.1 & 45k & 24.5 & 0.47 \\
    {\Large \color{colorsyn} $\bullet$} Syntactic Generic & 2615 & 36.2 & 65k & 31.6 & 0.40 \\
    \cmidrule(r){1-1} \cmidrule(lr){2-3}  \cmidrule(lr){4-5} \cmidrule(l){6-6}
    {\Large \color{colordis} $\bullet$} Reordering & 2379 & 34.6 & 75k & 32.2 & 0.50 \\
    {\Large \color{colordis} $\bullet$} Anaphora Resolut. & 302 & 5.4 & 13k & 7.6 & 0.30 \\
    {\Large \color{colordis} $\bullet$} Anaphora Insert. & 362 & 6.4 & 11k & 7.2 & 0.73 \\
    \cmidrule(r){1-1} \cmidrule(lr){2-3}  \cmidrule(lr){4-5} \cmidrule(l){6-6}
    {\Large \color{colorsem} $\bullet$} Elaboration - Bkgrd & 805 & 12.9 & 1164 & 0.7 & 0.18 \\
    {\Large \color{colorsem} $\bullet$} Elaboration - Exple & 139 & 2.4 & 139 & 0.1 & 0.05 \\
    {\Large \color{colorsem} $\bullet$} Elaboration - Generic & 3195 & 36.0 & 91k & 37.6 & 0.09 \\
    {\Large \color{colorsem} $\bullet$} Semantic Deletion & 12928 & 76.8 & 343k & 73.6 & 0.83 \\
    {\Large \color{colorsem} $\bullet$} Specific-to-General & 332 & 5.7 & 1227 & 0.8 & 0.25\\
    \cmidrule(r){1-1} \cmidrule(lr){2-3}  \cmidrule(lr){4-5} \cmidrule(l){6-6}
    {\Large \color{colornon} $\bullet$} Format & 2688 & 35.3 & 82k & 35.2 & 0.58 \\
    {\Large \color{colornon} $\bullet$} Noise Deletion & 693 & 10.6 & 14k & 7.9 & 0.58 \\
    {\Large \color{colornon} $\bullet$} Fact Correction & 290 & 5.0 & 4581 & 2.4 & 0.37 \\
    {\Large \color{colornon} $\bullet$} Extraneous Info & 3028 & 36.5 & 105k & 37.0 & 0.69 \\
    {\Large \color{colornon} $\bullet$} Miscellaneous & 241 & 3.6 & 1820 & 0.8 & 0.0 \\
    \bottomrule{}
    \end{tabular}
    }
    \caption{Edit categories in \dataset{}. For the manually and silver-annotated portions of the dataset, \textbf{N}: number of annotated instances, \textbf{\%$\exists$}: percentage of documents with edit, \textbf{$\kappa$} is Cohen's Kappa measuring inter-annotator agreement level}
    \label{table:edit_categories_extra}
\end{table}

Table~\ref{table:edit_categories_extra} summarizes additional statistics of \dataset{}. We find that the BIC model identifies edits at roughly the same rate as the manual annotation, with a few exceptions for long-tail categories such as Elaborations or Specific-to-Generic, this is due to low model recall on infrequent categories.

Overall, the class-level agreement level stands around 0.62, measured using Cohen's Kappa on 329 document pairs that were annotated by multiple editors. Table~\ref{table:edit_categories_extra} provides category-specific Cohen's Kappa, with the main trend showing higher agreement for frequent categories (Semantic Deletion, Sentence Split, Lexical), and lower agreement for infrequent categories. The agreement level is particularly low for elaboration categories, however, when merging the three categories of elaborations into a super-category, we measure an agreement level of 0.4, indicating that some agreement exists at a coarser level. Future work can therefore choose to combine the elaboration categories to remove disagreement from the annotations.

\section{Identification Models Supplemental}
\label{appendix:identification_extras}
This Section provides the additional content related to Section~\ref{section:identifiers} of the paper.

\subsection{Model Specifics}
\label{appendix:id_model_specifics}
We provide the implementation and training detail of each model included in the experiments of Section~\ref{section:identifiers}:

The \textbf{Category Classification (CC)} model, used in the Adjacent-CC, BI-CC, and Oracle-CC pipelined approaches is implemented as a finetuned RoBERTa-large model with a sequence classification head (i.e. a model that generates a single prediction for the entire sequence). The model was trained on a processed version of the training portion of \dataset{}, in which each document pair was leveraged to create several samples, each based on a single group in the annotations. For each new sample, an adjusted alignment sequence is created by reverting all edit operations that are not part of the sample's considered group. The model receives the adjusted alignment sequence and must predict the category of the represented edit. Crucially, the CC model is expecting to see a single category per input alignment sequence and does not consider overlapping and multi-category edits. The model we use in experiments was trained with a batch size of 16, Apex half-precision, for seven epochs at a learning rate of $10^{-5}$. The best checkpoint based on validation F-1 was selected, achieving a validation F-1 score of 77.5. We note that there's a crucial mismatch between train and prediction time in CC-based pipelines, as the CC model is trained on oracle groups, and at prediction time, certain configurations provide the model with imperfect groups (such as the Adjacent and BI groupers), which likely negatively affects performance. The training of the final model took roughly 1 hour on a single A100 GPU, and roughly 50 runs were conducted in iterations of model training.

The \textbf{BI} model, used in the grouping stage of the BI-CC model is a RoBERTa-large sequence tagging model that receives as input an alignment sequence and must predict for each edit operation whether the operation is at the beginning of (B) or inside (I) an edit group. We used an XML-like language to represent the alignment sequence for the model, using two operation starts (\texttt{<insert>} and \texttt{<delete>}) and two operation ends (\texttt{</insert>} and \texttt{</delete>}) which were added as special tokens to the model's vocabulary. The model was then trained to generate each operation's binary B/I tag at the corresponding beginning delimiter token. The model was trained using half-precision, and a learning rate of $10^{-5}$ for 10 epochs, selecting the model with the highest F-1 binary accuracy on the validation set of \dataset{}. The training of the final model took roughly 25 minutes on a single A100 GPU, and roughly 20 training runs were conducted in iterations of model training.

The \textbf{Category Tagging (CT)} model, used in the first stage of the CT-Single, CT-Adjacent, and CT-Rules models, follows a similar architecture as the BI model described above, but outputs one of the 19 simplification categories for each edit operation instead of a B/I indicator. Additionally, CT uses a \textit{multi-label} token-classification head to handle the case of multiple categories for an edit operation (e.g. for overlapping edit groups). For training, we used a batch size of 8 and a learning rate of $10^{-5}$ for 10 epochs. The final checkpoint was selected based on validation-set performance. The training of the final model took approximately 20 minutes on a single A100 GPU, and roughly 10 training runs were conducted in iterations of model training.

The \textbf{Rules} grouping method used in the second stage of the CT-Rules model, relied on category-specific statistics in the training portion of \dataset{}. Categories were split into two sub-groups: contiguous and global. For each category, we analyzed the percentage of annotated of edits of the given category that were contiguous (adjacent) in their operation group. For each edit category, if a majority of annotated cases were contiguous, the edit category was labeled as \textit{contiguous}, otherwise, it was labeled as \textit{global}. For categories marked as contiguous, the model generated groups for predicted operation types based on contiguous boundaries (identical to the Adjacent grouping method), and all operations of a given global category were organized into a single group.

The \textbf{BIC} model uses an identical model architecture to the CT model described above, but expands the label space from 19 category labels to 57 joint category-BI labels. Specifically, for each category label \texttt{<}\textit{cat}\texttt{>}, two additional labels are considered: \texttt{<}\textit{cat}\texttt{-B>} and \texttt{<}\textit{cat}\texttt{-I>}, indicating whether the operation is at the beginning or end of a group of this category. At training time, an edit operation is tagged with \texttt{<}\textit{cat}\texttt{>} if the category is present and additionally with either \texttt{<}\textit{cat}\texttt{-B>} or \texttt{<}\textit{cat}\texttt{-I>} according to the operation's position within the annotated group. At inference time, the model outputs one or more of the 57 joint labels at each edit operation’s start token. If \texttt{<}\textit{cat}\texttt{>} is predicted for a given category, then the associated BI label is chosen based on whether \texttt{}<\textit{cat}\texttt{-B>} or \texttt{<}\textit{cat}-I\texttt{>} has the higher predicted probability. For training, we used a batch size of 8 and a learning rate of $10^{-5}$ for 10 epochs. The model checkpoint was selected based on validation-set performance. The training of the final model took approximately 20 minutes on a single A100 GPU, and roughly 15 training runs were conducted in iterations of model training.

The \textbf{Seq2seq} model was implemented based on a BART-large model that we fine-tuned on a seq2seq task using an XML representation of the alignment sequence. Example processing of the illustrative Figure~\ref{fig:illustrative_example} would be:
\begin{quote}
    Input: ``The Mariinsky Theater is a <INS>very famous</INS> <DEL>historic</DEL> theater of opera and balet ...''
\end{quote}
\begin{quote}
    Output: ``The Mariinsky Theater is a <B;lexical>very famous</INS> <I;lexical>historic</DEL> theater of opera and balet ...''
\end{quote}
As illustrated in the example, the model was trained to replace generic operation beginning tags with a joint tag representing the category and the BI tag of the operation. The vocabulary of the model was expanded to include the 38 tokens representing all combinations of (category x (B,I)) tags. The model was trained on the preprocessed data following a standard sequence-to-sequence formulation, with a batch size of 6, a learning rate of $2*10^{-5}$, for ten epochs, and the model with the lowest validation loss was selected as a final model. Training of the final model required roughly one hour of training, and roughly 20 training runs were conducted in iterations of model training.

\subsection{BIC Performance Breakdown}
\label{appendix:bic_detail}

Table~\ref{table:bic_breakdown} reports the performance of the BIC model, individualized by category. We find that performance generally improves on categories as the number of examples in the dataset increases, giving evidence that further annotations of tail categories could lead to improved performance of the BIC model.

\begin{table}[]
    \resizebox{0.5\textwidth}{!}{%
    \begin{tabular}{lccccc}    
    \textbf{Edit Category} & \textbf{N} & \textbf{Cat F-1} & \textbf{\%Part} &  \textbf{\%Exact} \\
    \cmidrule(r){1-1} \cmidrule(lr){2-2}  \cmidrule(lr){3-5}
    {\Large \color{colorsem} $\bullet$} Semantic Deletion & 12928 & 87.8 & 73.0 & 76.3 \\
    {\Large \color{colorlex} $\bullet$} Lexical Edit & 6789 & 70.4 & 61.6 & 64.8 \\
    {\Large \color{colorsem} $\bullet$} Elaboration - Generic & 3195 & 40.8 & 34.9 & 35.1 \\
    {\Large \color{colornon} $\bullet$} Extraneous Info & 3028 & 75.3 & 47.7 & 55.0 \\
    {\Large \color{colorsyn} $\bullet$} Sentence Split & 3010 & 83.5 & 55.6 & 69.9 \\
    {\Large \color{colornon} $\bullet$} Format & 2688 & 73.3 & 60.5 & 65.6 \\
    {\Large \color{colorsyn} $\bullet$} Syntactic Generic & 2615 & 70.7 & 63.0 & 63.3 \\
    {\Large \color{colordis} $\bullet$} Reordering & 2379 & 51.1 & 27.1 & 51.1 \\
    {\Large \color{colorsyn} $\bullet$} Syntactic Deletion & 1889 & 54.0 & 47.9 & 47.9 \\
    {\Large \color{colorsem} $\bullet$} Elaboration - Bkgrd & 805 & 23.0 & 26.3 & 26.3 \\
    {\Large \color{colornon} $\bullet$} Noise Deletion & 693 & 61.1 & 48.7 & 48.7 \\
    {\Large \color{colordis} $\bullet$} Anaphora Insert. & 362 & 50.5 & 42.9 & 42.9 \\
    {\Large \color{colorlex} $\bullet$} Entity Edit & 359& 39.2 & 39.7 & 39.7 \\
    {\Large \color{colorsyn} $\bullet$} Sentence Fusion & 334 & 50.7 & 27.4 & 32.3 \\
    {\Large \color{colorsem} $\bullet$} Specific-to-General & 332 & 17.2 & 15.9 & 15.9 \\
    {\Large \color{colordis} $\bullet$} Anaphora Resolut. & 302 & 62.7 & 57.1 & 57.1 \\
    {\Large \color{colornon} $\bullet$} Miscellaneous & 241 & 45.2 & 28.9 & 31.6 \\
    {\Large \color{colornon} $\bullet$} Fact Correction & 290 & 47.7 & 31.8 & 40.9 \\
    {\Large \color{colorsem} $\bullet$} Elaboration - Exple & 139 & 11.1 & 16.7 & 16.7 \\
    \bottomrule{}
    \end{tabular}
    }
    \caption{Breakdown of BIC model per edit category. Categories are sorted in order of frequency in the dataset, and we report the three metrics that can be computed at the category level. Categories belong to five classes: {\Large \color{colorlex} $\bullet$} lexical, {\Large \color{colorsyn} $\bullet$} syntactic, {\Large \color{colordis} $\bullet$} discourse, {\Large \color{colorsem} $\bullet$} semantic, and {\Large \color{colornon} $\bullet$} non-simplification.} 
    \label{table:bic_breakdown}
\end{table}

\subsection{Out-of-domain identification performance \& model throughput}
\label{appendix:identification_ood}

Table~\ref{table:identification_ood} presents the results analogous to Table~\ref{table:identification_id} but for the out-of-domain test set. We do not observe a marked drop in performance, indicating that either the identification models are capable of generalizing to unseen Wikipedia categories, or that selected OOD categories are not truly out of distribution. We discuss the OOD test set selection further in the Limitations section.

We compute the throughput of each model to provide insights into the computational cost of identifying edits in document pairs. All models were benchmarked by the time they took to identify edits in the entire validation set (i.e., roughly 500 document pairs), using a single A-100 GPU on the same server, and we report normalized documents per second throughput (\textbf{Doc/s}). All models were tested at batch-size 1, which could disadvantage some neural methods. Results are summarized in the right-most column of Table~\ref{table:identification_ood}. We find that the BIC model is the second-fastest neural method behind CT-based models, confirming that joint modeling of the edit identification task positively affects both performance and efficiency.

\begin{table}[]
    \resizebox{0.5\textwidth}{!}{%
    \begin{tabular}{lccccc}
    \textbf{Model Name} & \textbf{Cat F1} & \textbf{Class F1} & \textbf{\%Part} & \textbf{\%Exact} & \textbf{Doc/s} \\
    \cmidrule(r){1-1} \cmidrule(r){2-3} \cmidrule(r){4-5} \cmidrule(r){6-6}
    Op Majority        & 36.5            & 40.1              & -                & -                & 2.7k               \\
    \cmidrule(r){1-1} \cmidrule(r){2-3} \cmidrule(r){4-5} \cmidrule(r){6-6}
    Adjacent-CC        & 59.5            & 61.7              & 43.5             & 46.5             & 3.4                \\
    BI-CC              & 67.8            & 69.4              & 54.1             & 57.6             & 2.5                \\
    Oracle-CC          & 83.5            & 85.2              & -                & -                & 2.7                \\
    \cmidrule(r){1-1} \cmidrule(r){2-3} \cmidrule(r){4-5} \cmidrule(r){6-6}
    CT-Single          & 73.5            & 76.3              & 28.4             & 28.4             & \textbf{23.3}       \\
    CT-Adjacent        & 73.5            & 76.3              & 58.6             & 61.8             & 23.2               \\
    CT-Rules           & 73.5            & 76.3              & 55.9             & 59.6             & 23.2               \\
    \cmidrule(r){1-1} \cmidrule(r){2-3} \cmidrule(r){4-5} \cmidrule(r){6-6}
    BIC                & \textbf{74.9}   & \textbf{76.6}     & \textbf{57.3}    & \textbf{62.1}    & 18.9               \\
    Seq2Seq            & 44.6            & 47.2              & 30.7             & 34.4             & 0.1                \\
    \bottomrule
    \end{tabular}
    }
    \caption{Out-of-domain test set edit identification results. \textbf{Doc/s} reports the throughput of each model in documents per second.}
    \label{table:identification_ood}
\end{table}

\section{Generation Models Supplemental}
\label{appendix:genertion_extras}

This Section provides the additional content related to Section~\ref{section:generators} of the paper.

\subsection{Model Specifics}

The \textbf{ACCESS} model was implemented using the original author's public code release\footnote{\url{https://github.com/facebookresearch/access}}, and the default conditioning parameters of 0.95 for length target, 0.75 for Levenshtein target, and 0.75 for the word-rank target.

The \textbf{Keep-it-Simple} model was implemented using the original author's public model release on the HuggingFace model hub\footnote{\url{https://huggingface.co/philippelaban/keep_it_simple}}. As recommended by the authors, we used a beam search (beam size of 4) to generate candidates, selecting the beam with the highest likelihood as the final generated candidate.

The \textbf{BART-\dataset{}} and \textbf{BART-\dataset{}-C} models were trained on the standard and cleaned versions of the \dataset{} dataset, using a standard sequence-to-sequence framing, in which the model received the original document as an input, and was trained to generate the simplified document. We trained the models with a learning rate of $2*10^{-5}$, a batch size of six for three epochs, and selected the final checkpoint based on validation loss, which reached 1.12 for \textbf{BART-\dataset{}} and 0.78 for \textbf{BART-\dataset{}-C}. Training required 6-10 hours for each model, on a single A-100 GPU, and 5 runs were completed in the development of the models. At generation time, we used beam search (beam size of 4) to generate candidate simplifications.

The \textbf{GPT3-davinci-003} model was implemented using OpenAI's API access to the GPT3 model, with the following prompt: ``Simplify the document below so it is accessible to a wider audience. Start of document:'', with newlines inserted to delimit the task definition, the document, and the expected output. We used default generation parameters provided in the interface, and estimate the cost of generation at \$10 for the 500 documents in the validation set. We note that it is unclear whether GPT3 qualifies as a zero-shot model for simplification, since it is trained on Wikipedia (amongst others), and has therefore been trained on a super-set of the data in \dataset{}, although it has not seen the explicit revision pairing available in \dataset{}.

\subsection{Example Generations}
\label{appendix:example_generated_simplifications}

In Tables~\ref{table:example_simplifications_part1}-\ref{table:example_simplifications_part2}, we provide the revision of the Wikipedia page about the ``Millimeter'', included in the validation set of \dataset{}. The Tables then provide the alignment sequence of six candidate simplifications: the human-written reference in Simple English Wikipedia, and the outputs of the ACCESS, Keep it Simple, BART-\dataset{}, BART-\dataset{}-C and GPT3-davinci-003 models.

\begin{table*}[]
    \centering
    \begin{tabular}{p{16cm}}
    \toprule
    \hspace{3.3cm} \Large{Complex Document -- English Wikipedia} \TBstrut \\
    \hline
    \TBstrut
    The micrometre (International spelling as used by the International Bureau of Weights and Measures; SI symbol: $\mu$m) or micrometer (American spelling), also commonly known as a micron, is an SI derived unit of length equaling 1x10-6 of a metre (SI standard prefix ``micro-'' = 10-6); that is, one millionth of a metre (or one thousandth of a millimetre, 0.001 mm, or about 0.000039 inch).
    The micrometre is a common unit of measurement for wavelengths of infrared radiation as well as sizes of biological cells and bacteria, and for grading wool by the diameter of the fibres. The width of a single human hair ranges from approximately 10 to 200 $\mu$m. The first and longest human chromosome is approximately 10$\mu$m in length.
    \TBstrut \\
    \hline
    \hspace{3.35cm} \Large{Reference -- Simple English Wikipedia} \TBstrut \\
    \hline
    \TBstrut 
    \hlred{The} \hlblue{A} micrometre \hlred{(International} \hlblue{(its American} spelling \hlred{as used by the International Bureau of Weights and Measures} \hlblue{is micrometer} ; \hlred{SI} symbol \hlred{: $\mu$m} \hlblue{is $\mu$m} ) \hlred{or micrometer(American spelling), also commonly known as} \hlblue{is} a \hlred{micron, is an SI derived} unit of length \hlred{equaling 1x10-6 of} \hlblue{in the SI measurement system. It can also be called} a \hlred{metre(SI standard prefix" micro-" = 10-6); that} \hlblue{micron. It} is \hlred{,} one millionth of a metre \hlred{(or one thousandth of a millimetre, 0.001 mm, or about 0.000039 inch)}. \hlred{The micrometre is a common unit of measurement for wavelengths of infrared radiation as well as sizes of biological cells and bacteria, and for grading wool by the diameter of the fibres. The width of a single human hair ranges from approximately 10 to 200 $\mu$m. The first and longest human chromosome is approximately 10$\mu$m in length.}
    \TBstrut \\
    \hline
    \hspace{6.35cm} \Large{ACCESS} \TBstrut \\
    \hline
    \TBstrut
    The micrometre(International spelling as used by the International Bureau of Weights and Measures; SI symbol: \hlred{$\mu$m} \hlblue{mm} ) or micrometer(American spelling), also commonly known as a micron, is an SI derived unit of length equaling \hlred{1x10-6} \hlblue{1x10 - 6} of a metre(SI standard \hlred{prefix" micro-" = 10-6} ) \hlred{;} \hlblue{,} that is, one millionth of a \hlred{metre(or one thousandth of a} millimetre, \hlred{0.001 mm,} \hlblue{01 mm} or \hlred{about 0.000039 inch)} \hlblue{about00039}. The micrometre is a common unit of measurement for wavelengths of infrared radiation as well as sizes of biological cells and bacteria \hlred{, and for grading wool by the diameter of the fibres}. The width of a single human hair \hlred{ranges} \hlblue{is found} from \hlred{approximately} \hlblue{about} 10 to \hlred{200 $\mu$m} \hlblue{200 mm}. The first and longest human chromosome is \hlred{approximately 10$\mu$m} \hlblue{about 10mm} in length. \TBstrut \\
    \hline
    \hspace{6cm} \Large{Keep it Simple} \TBstrut \\
    \hline
    \TBstrut The micrometre \hlred{(International} \hlblue{is a common} spelling \hlred{as} used by the International Bureau of Weights and Measures \hlred{;} \hlblue{. It is also used by the} SI \hlred{symbol: $\mu$m) or micrometer(American spelling)} , \hlred{also commonly known as a} \hlblue{which is used by all of the measures in the same way, and so is the} micron, \hlblue{which} is \hlred{an SI derived unit of length equaling} \hlblue{equivalent to} 1x10-6 of a metre \hlred{(SI standard prefix" micro-" = 10-6); that is, one millionth of a metre(or one thousandth of a millimetre, 0.001 mm, or about 0.000039 inch)}. \hlred{The micrometre is a common unit of measurement for wavelengths of infrared radiation as well as sizes of biological cells and bacteria, and for grading wool by the diameter of the fibres. The width of a single human hair ranges from approximately 10 to 200 $\mu$m. The first and longest human chromosome is approximately 10$\mu$m in length.} \TBstrut \\

    \bottomrule
    \end{tabular}
    \caption{``Millimeter'' Wikipedia page, reference simplification from SEW, and model outputs from the ACCESS and Keep it Simple models. Deletions are in \hlred{red} and insertions in \hlblue{blue}. See Table~\ref{table:example_simplifications_part2} for other model outputs.}
    \label{table:example_simplifications_part1}
\end{table*}

\begin{table*}[]
\centering
\begin{tabular}{p{16cm}}
        \toprule
        \hspace{6.0cm} \Large{BART-WikiLarge} \TBstrut \\
        \hline
        \TBstrut
        The micrometre \hlred{(International spelling as used by the International Bureau of Weights and Measures; SI symbol: $\mu$m) or micrometer(American spelling), also commonly} \hlblue{(also} known as \hlblue{the micrometer) is}
        a \hlred{micron, is an SI derived} unit of length \hlred{equaling 1x10-6} \hlblue{that is less than 1/10,000} of a metre \hlred{(SI standard prefix" micro-" = 10-6); that is, one millionth of a metre(or one thousandth of a millimetre, 0.001 mm, or about 0.000039 inch)}.
        The \hlred{micrometre} \hlblue{micrometer(I)} is \hlred{a common} \hlblue{the smallest} unit of measurement for \hlred{wavelengths} \hlblue{the number} of \hlred{infrared radiation as well as sizes} \hlblue{different types} of \hlred{biological cells and bacteria} \hlblue{microorganisms}, and for \hlred{grading wool by the diameter} \hlblue{different sizes} of \hlred{the fibres} \hlblue{cells}. The \hlred{width} \hlblue{length} of a \hlred{single human} hair \hlred{ranges} \hlblue{range varies} from \hlred{approximately 10} \hlblue{5} to \hlred{200 $\mu$m} \hlblue{20 cm(1 to 2 in)}. The first and longest \hlred{human} chromosome is \hlred{approximately 10$\mu$m} \hlblue{4.7 in length, the second is 6.5} in length.

        \TBstrut \\
        \hline
        \hspace{6.0cm} \Large{BART-\dataset{}} \TBstrut \\
        \hline
        \TBstrut
        The micrometre \hlred{(International spelling as used by the International Bureau of Weights and Measures; SI} \hlblue{(SI} symbol: $\mu$m) or micrometer(American spelling), also commonly known as a micron, is an SI \hlred{derived} unit of length \hlred{equaling 1x10-6 of a metre(SI standard prefix" micro-" = 10-6); that} \hlblue{. It} is \hlred{,} \hlblue{equal to} one millionth of a metre(or \hlred{one thousandth} \hlblue{1/1000} of a \hlred{millimetre} \hlblue{millimeter} , 0.001 mm, or about 0.000039 inch). The \hlred{micrometre is a common unit of measurement for wavelengths of infrared radiation as well as sizes of biological cells and bacteria, and for grading wool by the diameter of the fibres. The} width of a single human hair \hlred{ranges from approximately} \hlblue{is about} 10 to 200 $\mu$m. The first and longest human chromosome is approximately 10$\mu$m in length.
        \TBstrut \\
        \hline
        \hspace{5.9cm} \Large{BART-\dataset{}-C} \TBstrut \\
        \hline
        \TBstrut
        \hlred{The} \hlblue{A} micrometre \hlred{(International spelling as used by the International Bureau of Weights and Measures; SI symbol: $\mu$m) or micrometer(American spelling), also commonly} \hlblue{(also} known as \hlred{a} micron, is an SI derived unit of length \hlred{equaling 1x10-6 of a metre(SI standard prefix" micro-" = 10-6); that is,} \hlblue{equal to} one millionth of a metre \hlred{(or one thousandth of a millimetre, 0.001 mm, or about 0.000039 inch)}. \hlred{The micrometre} \hlblue{It} is a common unit of measurement \hlred{for} \hlblue{of} wavelengths of infrared radiation as well as sizes of biological cells and bacteria, and for grading wool by the diameter of the fibres. The width of a single human hair ranges from approximately 10 to 200 $\mu$m. The first and longest human chromosome is approximately 10$\mu$m in length.
        \TBstrut \\
        \hline
        \hspace{5.8cm} \Large{GPT3-davinci-003} \TBstrut \\
        \hline
        \TBstrut
        \hlred{The} \hlblue{A} micrometre \hlred{(International spelling as used by the International Bureau of Weights and Measures; SI symbol} \hlblue{(symbol} : $\mu$m) \hlred{or micrometer(American spelling), also commonly known as} \hlblue{is} a \hlred{micron, is an SI derived unit of length equaling 1x10-6 of a metre(SI standard prefix" micro-" = 10-6); that is, one} millionth of a metre \hlred{(or} \hlblue{, or} one thousandth of a millimetre \hlred{, 0.001 mm, or about 0.000039 inch} \hlblue{(0.001 mm} ). \hlred{The micrometre is a common unit of measurement for wavelengths} \hlblue{It's used to measure the wavelength} of infrared radiation \hlred{as well as sizes} \hlblue{, the size} of biological cells and bacteria, and \hlred{for grading wool by} the diameter of \hlred{the} \hlblue{wool} fibres. \hlred{The width of a} \hlblue{A} single human hair \hlred{ranges from approximately 10 to 200 $\mu$m. The first} \hlblue{is around 10-200 $\mu$m wide,} and \hlblue{the} longest human chromosome is \hlred{approximately 10$\mu$m} \hlblue{about 10 $\mu$m} in length.
        \TBstrut \\
        \bottomrule
\end{tabular}
\caption{Model outputs from the two \dataset{}-trained models and the GPT3 zero-shot baseline. Deletions are in \hlred{red} and insertions in \hlblue{blue}.}
\label{table:example_simplifications_part2}
\end{table*}

\end{document}